\documentclass[sw,crcready]{iosart2x}

\usepackage[dvipsnames]{xcolor}

\usepackage{tikz}
\usetikzlibrary{arrows,shapes,positioning,shadows,trees}

\definecolor{myred}{HTML}{EA6B66}
\definecolor{mygrey}{HTML}{999999}

\tikzset{
  basic/.style  = {draw, text width=2cm, drop shadow, font=\sffamily, rectangle},
  root/.style   = {basic, rounded corners=2pt, thin, align=center,fill=white, text=myred},
  level 2/.style = {basic, rounded corners=6pt, thin,align=center, fill=white,text width=10em, text=myred},
  level 3/.style = {thin, align=left, fill=white, text width=7.8em}
}

\usepackage{hyperref}

\usepackage[strings]{underscore}

\usepackage[flushleft]{threeparttable}
\usepackage{tablefootnote}

\usepackage{amsthm}
\theoremstyle{definition}

\usepackage{pifont}
\usepackage{comment}
\usepackage{enumitem}
\usepackage{booktabs}
\usepackage{pifont}

\usepackage{setspace}
\usepackage{array}

\renewcommand{\textsf}[1]{{\small\sffamily#1}}

\newcommand{\multirowcell}[1]{\setlength\extrarowheight{-5pt}\begin{tabular}[c]{@{}c@{}}#1\end{tabular}}

\newcommand{\mycite}[1]{\citeauthor{#1}~(\citeyear{#1})~\cite{#1}}

\newcommand\copyrighttext{%
  \scriptsize \textcolor{blue}{Disclaimer: \copyright \, \"Ozge Sevgili, Artem Shelmanov, Mikhail Arkhipov, Alexander Panchenko, and Chris Biemann, 2022. The definitive, peer reviewed and edited version of this article is published in the Semantic Web Journal, Special Issue on Deep Learning and Knowledge Graphs, 2022}
}

\pubyear{0000}
\volume{0}
\firstpage{1}
\lastpage{1}

\begin{document}

\begin{frontmatter}

\copyrighttext 

\title{Neural Entity Linking: A Survey of Models Based on Deep Learning}
\runtitle{Neural Entity Linking: A Survey of Models Based on Deep Learning}



\author[A]{\inits{\"{O}.}\fnms{\"{O}zge} \snm{Sevgili}\ead[label=e1]{oezge.sevgili.ergueven@studium.uni-hamburg.de}%
\thanks{Equal contribution. Corresponding author. \printead{e1}.}%
},
\author[D,B,C]{\inits{A.}\fnms{Artem} \snm{Shelmanov}\ead[label=e2]{shelmanov@airi.net}
 \thanks{Equal contribution. Corresponding author. \printead{e2}.}
},
\author[E]{\inits{M.}\fnms{Mikhail} \snm{Arkhipov}\ead[label=e3]{arkhipov@yahoo.com}},
\author[B]{\inits{A.}\fnms{Alexander} \snm{Panchenko}\ead[label=e4]{a.panchenko@skoltech.ru}},
\author[A]{\inits{C.}\fnms{Chris} \snm{Biemann}\ead[label=e5]{christian.biemann@uni-hamburg.de}}

\address[A]{Language Technology Group, \orgname{Universit{\"a}t Hamburg, Informatikum, Vogt-K\"{o}lln-Stra\ss{}e 30, 22527 Hamburg},
 \cny{Germany}\printead[presep={\\}]{e1,e5}}
\address[B]{Center for Artificial Intelligence Technologies, \orgname{Skolkovo Institute of Science and Technology, Bolshoy Boulevard 30, bld. 1, 121205, Moscow}, \cny{Russia}\printead[presep={\\}]{e4}}
\address[C]{Research Computing Center, \orgname{Lomonosov Moscow State University, GSP-1, Leninskie Gory, 119991, Moscow},  \cny{Russia}}
\address[D] {AIRI, Nizhny Susalny lane 5 p. 19, 105064, Moscow, \cny{Russia}\printead[presep={\\}]{e2}}
\address[E]{Neural Networks and Deep Learning Laboratory, \orgname{Moscow Institute of Physics and Technology, 9 Institutskiy per., Dolgoprudny, 141701, Moscow},  \cny{Russia}\printead[presep={\\}]{e3}}

\begin{review}{editor}
\reviewer{\fnms{Mehwish} \snm{Alam}\address{\orgname{FIZ Karlsruhe - Leibniz Institute for Information Infrastructure}, \cny{Germany}}}
\reviewer{\fnms{Davide} \snm{Buscaldi}\address{\orgname{LIPN, Université Sorbonne Paris Nord}, \cny{France}}}
\reviewer{\fnms{Michael} \snm{Cochez}\address{\orgname{Vrije University of Amsterdam}, \cny{the Netherlands}}}
\reviewer{\fnms{Francesco} \snm{Osborne}\address{\orgname{Knowledge Media Institute, (KMi), The Open University}, \cny{UK}}}
\reviewer{\fnms{Diego Reforgiato} \snm{Recupero}\address{\orgname{University of Cagliari}, \cny{Italy}}}
\reviewer{\fnms{Harald} \snm{Sack}\address{\orgname{FIZ Karlsruhe - Leibniz Institute for Information Infrastructure}, \cny{Germany}}}
\end{review}

\begin{review}{solicited}
\reviewer{\fnms{Italo Lopes} \snm{Oliveira}\address{\orgname{University or Company name}, \cny{Country}}}
\reviewer{\fnms{Sahar} \snm{Vahdati}\address{\orgname{University or Company name}, \cny{Country}}}
\reviewer{\fnms{Mojtaba} \snm{Nayyeri}\address{\orgname{University or Company name}, \cny{Country}}}
\reviewer{\fnms{Daza} \snm{Cruz}\address{\orgname{University or Company name}, \cny{Country}}}
\reviewer{\fnms{Anonymous} \snm{}\address{\orgname{University or Company name}, \cny{Country}}}
\end{review}
\begin{review}{open}
\reviewer{\fnms{First Open} \snm{Reviewer}\address{\orgname{University or Company name}, \cny{Country}}}
\reviewer{\fnms{Second Open} \snm{Reviewer}\address{\orgname{University or Company name}, \cny{Country}}}
\end{review}

\begin{abstract}
This survey presents a comprehensive description of recent neural entity linking (EL) systems developed since 2015 as a result of the ``deep learning revolution'' in natural language processing. Its goal is to systemize design features of neural entity linking systems and compare their performance to the remarkable classic methods on common benchmarks. This work distills a generic architecture of a neural EL system and discusses its components, such as candidate generation, mention-context encoding, and entity ranking, summarizing prominent methods for each of them. The vast variety of modifications of this general architecture are grouped by several common themes: joint entity mention detection and disambiguation, models for global linking, domain-independent techniques including zero-shot and distant supervision methods, and cross-lingual approaches. Since many neural models take advantage of entity and mention/context embeddings to represent their meaning, this work also overviews prominent entity embedding techniques. Finally, the survey touches on applications of entity linking, focusing on the recently emerged use-case of enhancing deep pre-trained masked language models based on the Transformer architecture. 
\end{abstract}


\end{frontmatter}




\section{Introduction}
\label{intro}
%
%

    


Knowledge Graphs (KGs), such as Freebase \cite{freebase}, DBpedia \cite{dbpedia}, and Wikidata~\cite{wikidata}, contain rich and precise information about entities of all kinds, such as persons, locations, organizations, movies, and scientific theories, just to name a few. Each entity has a set of carefully defined relations and attributes, e.g. ``was born in'' or ``play for''. This wealth of structured information gives rise to and facilitates the development of semantic processing algorithms as they can directly operate on and benefit from such entity representations. For instance, imagine a search engine that is able to retrieve mentions in the news during the last month of all retired NBA players with a net income of more than 1 billion US dollars. The list of players together with their income and retirement information may be available in a knowledge graph. Equipped with this information, it appears to be straightforward to look up mentions of retired basketball players in the newswire. However, the main obstacle in this setup is the lexical ambiguity of entities. In the context of this application, one would want to only retrieve all mentions of ``Michael Jordan (basketball player)''\footnote{\url{https://en.wikipedia.org/wiki/Michael_Jordan}} and exclude mentions of other persons with the same name such as ``Michael Jordan (mathematician)''\footnote{\url{https://en.wikipedia.org/wiki/Michael_I._Jordan}}. 

This is why Entity Linking (EL) -- the process of matching a mention, e.g. ``Michael Jordan'', in a textual context to a KG record (e.g. ``basketball player'' or ``mathematician'') fitting the context -- is the key technology enabling various semantic applications. Thus, EL is the task of identifying an entity mention in the (unstructured) text and establishing a link to an entry in a (structured) knowledge graph. 

Entity linking is an essential component of many information extraction (IE) and natural language understanding (NLU) pipelines since it resolves the lexical ambiguity of entity mentions and determines their meanings in context.
A link between a textual mention and an entity in a knowledge graph also allows us to take advantage of the information encompassed in a semantic graph, which is shown to be useful in such NLU tasks as information extraction, biomedical text processing, or semantic parsing and question answering (see Section \ref{sec:applications}). This wide range of direct applications is the reason why entity linking is enjoying great interest from both academy and industry for more than two decades.

\subsection{Goal and Scope of this Survey}

Recently, a new generation of approaches for entity linking based on neural models and deep learning emerged, pushing the state-of-the-art performance in this task to a new level. The goal of our survey is to provide an overview of this latest wave of models, emerging from 2015.
 
Models based on neural networks have managed to excel in EL as in many other natural language processing tasks due to their ability to learn useful distributed semantic representations of linguistic data~\cite{collobert2011natural,young2018recent,bengio}. These current state-of-the-art neural entity linking models have shown significant improvements over ``classical''\footnote{On classical ML vs deep learning: \url{https://towardsdatascience.com/deep-learning-vs-classical-machine-learning-9a42c6d48aa}} machine learning approaches \cite{lazic,ratinov,chisholm} to name a few
that are based on shallow architectures, e.g. Support Vector Machines, and/or depend mostly on hand-crafted features. Such models often cannot capture all relevant statistical dependencies and interactions \cite{hofmann}. In contrast, deep neural networks are able to learn sophisticated representations within their deep layered architectures. This reduces the burden of manual feature engineering and enables significant improvements in EL and other tasks.

In this survey, we systemize recently proposed neural models, distilling one generic architecture used by the majority of neural EL models (illustrated in Figures \ref{fig:edpipeline} and \ref{fig:ranking}). We describe the models used in each component of this architecture, e.g. candidate generation, mention-context encoding, entity ranking. Prominent variations of this generic architecture, e.g. end-to-end EL or global models, are also discussed. To better structure the sheer amount of available models, various types of methods are illustrated in taxonomies (Figures~\ref{fig:taxonomypipelinetikz} and \ref{fig:taxonomymodificationstikz}), while notable features of each model are carefully assembled in a tabular form (Table \ref{table:elmodels}). We discuss the performance of the models on commonly used entity linking/disambiguation benchmarks and an entity relatedness dataset. Because of the sheer amount of work, it was not possible for us to try available software and to compare approaches on further parameters, such as computational complexity, run-time, and memory requirements. Nevertheless, we created a comprehensive collection of references to publicly available official implementations of EL models and systems discussed in this survey (see Table \ref{table:sourcecodes} in Appendix A).

An important component of neural entity linking systems is distributed entity representations and entity encoding methods. It has been shown that encoding the KG structure (entity relationships), entity definitions, or word/entity co-occurrence statistics from large textual corpora in low-dimensional vectors improves the generalization capabilities of EL models \cite{huang,hofmann}. Therefore, we also summarize distributed entity representation models and novel methods for entity encoding.

Many natural language processing systems take advantage of deep pre-trained language models like ELMo \cite{peters2018deep}, BERT \cite{bert}, and their modifications. EL made its path into these models as a way of introducing information stored in KGs, which helps to adapt word representations to some text processing tasks. We discuss this novel application of EL and its further development.

\subsection{Article Collection Methodology} \label{subsec:surveymethodology}

We do not have a strict article collection algorithm for the review like e.g., the one conducted by \citet{holisticsurvey}. Our main goal is to provide and describe a conceptual framework that can be applied to the majority of recently presented neural approaches to EL. Nevertheless, as with all surveys, we had to draw the line somewhere. The main criteria for including papers into this survey was that they had been published during or after 2015, and they primarily address the task of EL, i.e. resolving textual mentions to entries in KGs, or discussing EL applications. We explicitly exclude related work e.g., on (fine-grained) entity typing (see \cite{aly-etal-2021-leveraging,openentity}), which also encompasses a disambiguation task, and work that employs KGs for other tasks than EL. This survey also does not try to cover all EL methods designed for specific domains like biomedical texts or messages in social media. For the general-purpose EL models evaluated on well-established benchmarks, we try to be as comprehensive as possible with respect to recent-enough papers that fit into the conceptual framework, no matter where they have appeared (however, with a focus on top conferences and journals in the fields of natural language processing and Semantic Web).



\begin{figure*}
    \centering
    \includegraphics[width=1.0\textwidth]{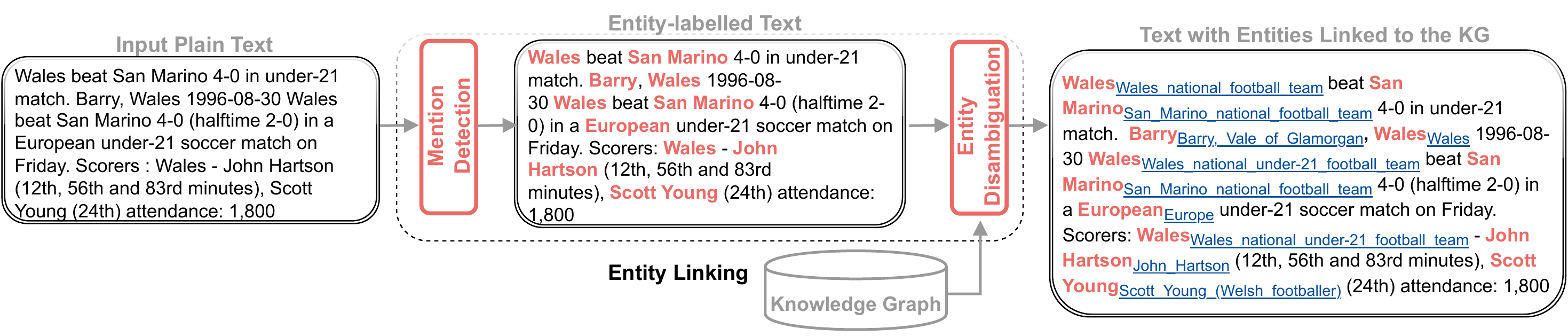}
    \caption{\textbf{The entity linking task}. An Entity Linking (EL) model takes a raw textual input and enriches it with entity mentions linked to nodes in a Knowledge Graph (KG). The task is commonly split into entity mention detection and entity disambiguation sub-tasks. 
    }
    \label{fig:systemdesc}
\end{figure*}

\subsection{Previous Surveys}

One of the first surveys on EL was prepared by \citet{surveywei} in 2015. They cover the main approaches to entity linking (within the modules, e.g. candidate generation, ranking), its applications, evaluation methods, and future directions. In the same year, \citet{surveyling} presented a work that aims to provide (1) a standard problem definition to reduce confusion that appears due to the existence of variant similar tasks related to EL (e.g., Wikification \cite{milne} and named entity linking \cite{hoffart}), and (2) a clear comparison of models and their various aspects.

There are also other surveys that address a wider scope. The work of \citet{surveyswj}, published in 2020,
 involves information extraction models and semantic web technologies. 
Namely, they consider many tasks, like named entity recognition, entity linking, terminology extraction, keyphrase extraction, topic modeling, topic labeling, relation extraction tasks.
In a similar vein, the work of \citet{ieeeelsurvey}, released in 2020, overviews the research in named entity recognition, named entity disambiguation, and entity linking published between 2014 and 2019. 

Another recent survey paper by \citet{holisticsurvey}, published in 2020, analyses and summarizes EL approaches that exhibit some holism. This viewpoint limits the survey to the works that exploit various peculiarities of the EL task: additional metadata stored in specific input like microblogs, specific features that can be extracted from this input like geographic coordinates in tweets, timestamps, interests of users posted these tweets, and specific disambiguation methods that take advantage of these additional features. In the concurrent work, \citet{moller2021survey} overview models developed specifically for linking English entities to the Wikidata \cite{wikidata} and discuss features of this KG that can be exploited for increasing the linking performance. 



Previous surveys on similar topics (a) do not cover many recent publications \cite{surveyling,surveywei}, (b) broadly cover numerous topics \cite{surveyswj,ieeeelsurvey}, or (c) are focused on the specific types of methods \cite{holisticsurvey} or a knowledge graph \cite{moller2021survey}. There is not yet, to our knowledge, a detailed survey specifically devoted to recent neural entity linking models. The previous surveys also do not address the topics of entity and context/mention encoding, applications of EL to deep pre-trained language models, and cross-lingual EL. We are also the first to summarize the domain-independent approaches to EL, several of which are based on zero-shot techniques. 


\subsection{Contributions}

More specifically, this article makes the following contributions: 
\begin{itemize}[itemsep=1mm, parsep=0pt]
   \item a survey of state-of-the-art neural entity linking models;
   \item a systematization of various features of neural EL methods and their evaluation results on popular benchmarks;
   \item a summary of entity and context/mention embedding techniques;
   \item a discussion of recent domain-independent (zero-shot) and cross-lingual EL approaches;
   \item a survey of EL applications to modeling word representations.
\end{itemize}

The structure of this survey is the following. We start with defining the EL task in Section \ref{sec:taskdesc}. In Section \ref{subsec:generalapproaches}, the general architecture of neural entity linking systems is presented. Modifications and variations of this basic pipeline are discussed in Section \ref{subsec:modifications}. In Section \ref{sec:evaluation}, we summarize the performance of EL models on standard benchmarks and present results of the entity relatedness evaluation. Section \ref{sec:applications} is dedicated to applications of EL with a focus on recently emerged applications for improving neural language models. Finally, Section \ref{sec:conclusion} concludes the survey and suggests promising directions of future work.


\begin{figure*}[t]
\centering
\includegraphics[width=1.0\textwidth]{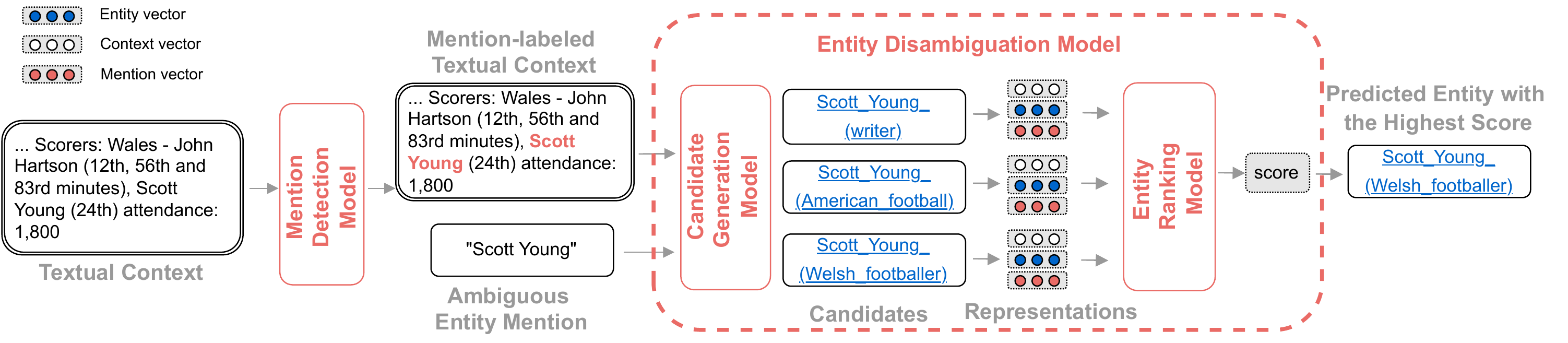}

\caption{\textbf{General architecture for neural entity linking.} Entity Linking (EL) consists of two main steps: \textit{Mention Detection (MD)}, when entity mention boundaries in a text are identified, and \textit{Entity Disambiguation (ED)}, when a corresponding entity is predicted for the given mention. Entity disambiguation is further carried out in two steps: \textit{Candidate Generation}, when possible candidate entities are selected for the mention, and \textit{Entity Ranking}, when a correspondence score between context/mention and each candidate is computed through the comparison of their vector representations.}

\label{fig:edpipeline}
\end{figure*}


\section{Task Description} \label{sec:taskdesc}

\subsection{Informal Definition}

Consider the example presented in Figure \ref{fig:systemdesc} with an entity mention \textit{Scott Young} in a soccer-game-related context. Literally, this common name can refer to at least three different people: the \textit{American football player}, the \textit{Welsh football player}, or the \textit{writer}. The EL task is to (1) correctly detect the entity mention in the text, (2) resolve its ambiguity and ultimately provide a link to a corresponding entity entry in a KG, e.g. provide for the \textit{Scott Young} mention in this context a link to the \textit{Welsh footballer}\footnote{\url{https://en.wikipedia.org/wiki/Scott_Young_(Welsh_footballer)}} instead of the \textit{writer}\footnote{\url{https://en.wikipedia.org/wiki/Scott_Young_(writer)}}. To achieve this goal, the task is usually decomposed into two sub-tasks, as illustrated in Figure~\ref{fig:systemdesc}: Mention Detection (MD) and Entity Disambiguation (ED).

\subsection{Formal Definition}

\subsubsection{Knowledge Graph (KG)}

A KG contains entities, relations, and facts, where facts are denoted as triples (i.e. head entity, relation, tail entity) as defined in \citet{kgsurvey}. Formally, as defined by~\citet{farber2018linked}, a KG is a set of RDF triples where each triple $(s,p,o)$ is an ordered set of the following terms: a subject $s \in U \cup B$, a predicate $p \in U$, and an object $o \in U \cup B \cup L$. An RDF term is either a URI $u \in U$, a blank node $b \in B$, or a literal $l \in L$. URI (or IRI) nodes are for the global identification of entities on the Web; literal nodes are for strings and other datatype values (e.g. integers, dates); and the blank node is for anonymous nodes, which are not assigned an identifier, as explained in \citet{hogan2021knowledge}.

This RDF representation can be considered as a multi-relational graph 
$G=(E,\mathbb{A}=\{A_0,A_1,...,A_m \subseteq
 (E \times E) \})$, where $E$ is a set of all entities of a KG, and  $\mathbb{A}$ is a family of typed edge sets of length $m$. For example, $A_0$ is the ``\textsf{occupation}'' predicate adjacency matrix, $A_1$ is the ``\textsf{founded}'' predicate adjacency matrix, etc.
 
 There is also an equivalent three-way tensor representation of a KG $\mathcal{A} \in \{0, 1\}^{n\times m \times n}$, where
\begin{equation}
\mathcal{A}_{i,k,j} = 
    \begin{cases}
     1 & \text{if $(i,j) \in A_k : k \leq m$} \\
     0 & \text{otherwise.}
   \end{cases}
\end{equation}

\subsubsection{Mention Detection (MD)}

The goal of mention detection is to identify an entity mention span, while entity disambiguation performs linking of found mentions to entries of a KG. We can consider this task as determining an $\mathsf{MD}$ function that takes as input a textual context $c_i \in C$ (e.g. a document in a document collection) and outputs a sequence of $n$ mentions $(m_1,\dots m_n)$ in this context $m_i \in M$, where $M$ is a set of all possible text spans in the context:
\begin{equation}
    \mathsf{MD} : C \xrightarrow{} M^n.
\label{eq:recognition}
\end{equation}

In the majority of works on EL, it is assumed that the mentions are already given or detected, for example, using a named entity recognition (NER) system (sometimes called named entity recognition and classification (NERC) \cite{aly-etal-2021-leveraging,nadeau2007survey}). We should note that, usually, in addition to MD, NER systems also tag/classify mentions with a predefined types \cite{nersurvey2022,rel,holisticsurvey,martins} that also can be leveraged for disambiguation \cite{martins}.

\subsubsection{Entity Disambiguation (ED)}

The entity disambiguation task can be considered as determining a function $\mathsf{ED}$ that, given a sequence of $n$ mentions in a document and their contexts $(c_1, \dots, c_n)$, outputs an entity assignment $(e_1, \dots,e_n), e_i \in E$, where $E$ is a set of entities in a KG:
\begin{equation}
    \mathsf{ED} : (M, C)^n \xrightarrow{} E^n.
\label{eq:disambiguator}
\end{equation}

To learn a mapping from entity mentions in a context to entity entries in a KG, EL models use supervision signals like manually annotated mention-entity pairs. The size of KGs varies; they can contain hundreds of thousands or even millions of entities. Due to their large size, training data for EL would be extremely unbalanced; training sets can lack even a single example for a particular entity or mention, e.g. as in the popular AIDA corpus \cite{hoffart}. To deal with this problem, EL models should have wide generalization capabilities. 

Despite KGs being usually large, they are incomplete. Therefore, some mentions in a text cannot be correctly mapped to any KG entry. Determining such unlinkable mentions, which usually is designated as linking to a $\mathsf{NIL}$ entry, is one of the current EL challenges. Methods that address this problem provide a separate function for it or extend the set of entities in the disambiguation function with this special entry:
\begin{equation}
    \mathsf{ED} : (M, C)^n \xrightarrow{} (E\cup \mathsf{NIL})^n.
\label{eq:disambiguator2}
\end{equation}

\subsection{Terminological Aspects}

More or less, the same technologies and models are sometimes called differently in the literature. Namely, Wikification \cite{cheng2013relational} and entity disambiguation are considered as subtypes of EL \cite{naviglielwsd}. To be comprehensive in this survey, we assume that the entity linking task encompasses both entity mention detection and entity disambiguation. However, only a few studies suggest models that perform MD and ED jointly, while the majority of papers on EL focus exclusively on ED and assume that mention boundaries are given by an external entity recognizer \cite{rizzo-etal-2014-benchmarking} (which may lead to some terminological confusions). Numerous techniques that perform MD (e.g. in the NER task) without entity disambiguation are considered in many previous surveys \cite{nadeau2007survey,sharnagat2014named,goyal2018recent,yadav-bethard-2018-survey,nersurvey2022} inter alia and are out of the scope of this work.

Entity linking in the general case is not restricted to linking mentions to graph nodes but rather to concepts in a knowledge base. However, most of the modern widely-used knowledge bases organize information in the form of a graph \cite{dbpedia, freebase, wikidata}, even in particular domains, like e.g. the scholarly domain \cite{kgforscholarlydomain}. A basic statement in a data/knowledge base usually can be represented as a subject-predicate-object tuple $(s,p,o)$, e.g. \textsf{(John\_Lennon, occupation, singer)} or \textsf{(New\_York\_City, founded, 1624)}, and a set of such tuples can be represented as a multi-relational graph. This formalism helps to efficiently organize knowledge for many applications ranging from search engines to question answering and recommendation systems \cite{hogan2021knowledge,kgsurvey}. Therefore, in this article, the terms Knowledge Graph (KG) and Knowledge Base (KB) are used interchangeably.

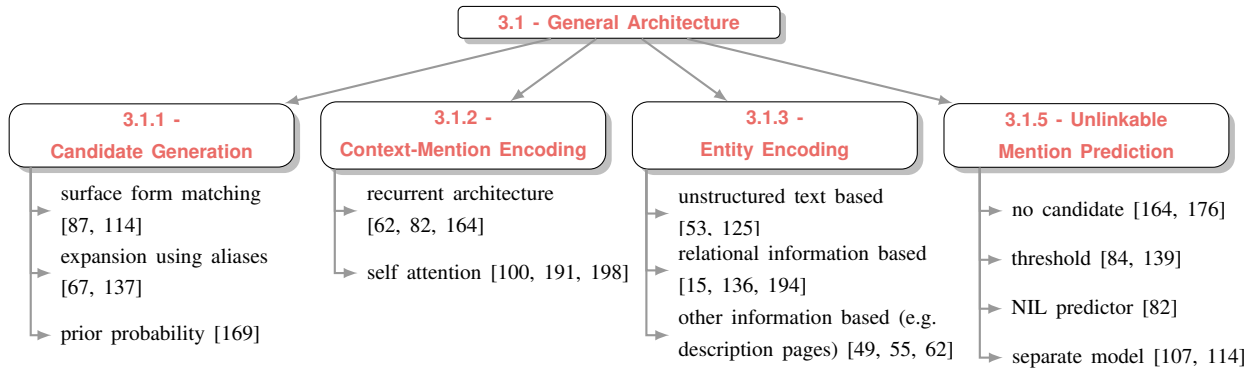
\begin{figure*}
    \centering
    \begin{tikzpicture}[
  level 1/.style={sibling distance=41mm},
  edge from parent/.style={->,draw, mygrey, thick},
  >=latex]

\node[root, text width=4cm] {\footnotesize \textbf{3.1 - General Architecture}}
  child {node[level 2] (c1) {\footnotesize \textbf{3.1.1 - \\ Candidate Generation}}}
  child {node[level 2] (c2) {\footnotesize \textbf{3.1.2 - 
  \\ Context-Mention Encoding}}}
  child {node[level 2] (c3) {\footnotesize \textbf{3.1.3 - \\ Entity Encoding}}}
  child {node[level 2] (c4) {\footnotesize \textbf{3.1.5 - Unlinkable Mention Prediction}}};

\begin{scope}[every node/.style={level 3}]
\node [below of = c1, xshift=5pt] (c11) {\footnotesize surface form matching  \cite{titov1,moreno}};
\node [below of = c11, yshift=5pt] (c12) {\footnotesize  expansion using aliases  \cite{hoffart,pershina}};
\node [below of = c12, yshift=5pt] (c13) {\footnotesize prior probability  \cite{crosswiki}};

\node [below of = c2, xshift=3pt] (c21) {\footnotesize recurrent architecture  \cite{gupta, end2end, sil}};
\node [below of = c21, yshift=5pt, xshift=25pt, text width=4.5cm] (c22) {\footnotesize self attention \cite{zero-shot, wufacebook, yamada20}};

\node [below of = c3, xshift=28pt, yshift=-1pt, text width=4.5cm] (c31)  {\footnotesize unstructured text based \\ \cite{hofmann,griffis}};
\node [below of = c31, yshift=7pt, text width=4.5cm] (c32) {\footnotesize relational information based \\ \cite{deepwalk,transe,yamada}};

\node [below of = c32, yshift=4pt, text width=4.5cm] (c33) {\footnotesize other information based (e.g. \\ description pages) \cite{gupta,francis2016capturing,gillick}};

\node [below of = c4, xshift=22pt, text width=3.5cm] (c41) {\footnotesize no candidate \cite{sil, crossling-wikification}};
\node [below of = c41, xshift=-11pt, yshift=10pt] (c42) {\footnotesize threshold \cite{lazic,knowbert}};
\node [below of = c42, yshift=10pt] (c43) {\footnotesize NIL predictor \cite{end2end}};
\node [below of = c43, yshift=10pt, xshift=11pt, text width=3.5cm] (c44) {\footnotesize separate model \cite{moreno,martins}};
\end{scope}

\foreach \value in {1,2,3}
  \draw[->, mygrey,thick] (c1.195) |- (c1\value.west);

\foreach \value in {1,2}
  \draw[->, mygrey,thick] (c2.195) |- (c2\value.west);

\foreach \value in {1,2,3}
  \draw[->, mygrey,thick] (c3.195) |- (c3\value.west);
\foreach \value in {1,2,3,4}
  \draw[->, mygrey,thick] (c4.197) |- (c4\value.west);
\end{tikzpicture}
    \caption{\textbf{Reference map of the general architecture of neural EL systems.} The categorization of each step in the general neural EL architecture with alternative design choices and example references illustrating each of the choices.}
    \label{fig:taxonomypipelinetikz}
\end{figure*}


\section{Neural Entity Linking} \label{sec:neuralentitylinking}

We start the discussion of neural entity linking approaches from the most general architecture of EL pipelines and continue with various specific modifications like joint entity mention detection and linking, disambiguation techniques that leverage global context, domain-independent EL approaches including zero-shot methods, and cross-lingual models.

\subsection{General Architecture} \label{subsec:generalapproaches}

Some of the attempts to EL based on neural networks treat it as a multi-class classification task in which entities correspond to classes. However, the straightforward approach results in a large number of classes, which leads to suboptimal performance without task-sharing \cite{kar}. The streamlined approach to EL is to treat it as a ranking problem. We present the generalized EL architecture in Figure \ref{fig:edpipeline}, which is applicable to the majority of neural approaches. Here, the mention detection model identifies the mention boundaries in text. The next step is to produce a shortlist of possible entities (candidates) for the mention, e.g. producing \textsf{Scott\_Young\_(writer)} as a candidate rather than a completely random entity. Then, the mention encoder produces a semantic vector representation of a mention in a context. The entity encoder produces a set of vector representations of candidates. Finally, the entity ranking model compares mention and entity representations and estimates mention-entity correspondence scores. An optional step is to determine unlinkable mentions, for which a KG does not contain a corresponding entity. The categorization of each step in the general neural EL architecture is summarized in Figure \ref{fig:taxonomypipelinetikz}.


\begin{table*}
\footnotesize
\centering
\caption{\textbf{Candidate generation examples.} Candidate entities for the example mention ``Big Blue'' obtained using several candidate generation methods. The highlighted candidates are ``correct'' entities assuming that the given mention refers to the IBM corporation and not a river, e.g. {\footnotesize \textsf{Big\_Blue\_River\_(Kansas)}}. }
\scalebox{1.0}{

\begin{tabular}{c|c} \bf Method & \bf 5 candidate entities for the example mention ``Big Blue'' \\ \hline 

\multirowcell{\bf surface form matching based\\ on DBpedia names\protect \footnotemark} & 

\multirowcell{{ \textsf{Big\_Blue\_Trail}}, \textsf{Big\_Bluegrass}, \textsf{Big\_Blue\_Spring\_cave\_crayfish},\\ \textsf{ Dexter\_Bexley\_and\_the\_Big\_Blue\_Beastie}, \textsf{IBM\_Big\_Blue\_(X-League)}}
\\ [0.2cm] \hline

\multirowcell{\bf expansion using aliases \\ from YAGO-means\protect\footnotemark} & \multirowcell{\textsf{ Big\_Blue\_River\_(Indiana)}, \textsf{ Big\_Blue\_River\_(Kansas)},\\
\textsf{ Big\_Blue\_(crane)}, \textsf{Big_Red_(drink)}, \textsf{\textbf{IBM}}}
\\ [0.2cm] \hline

\multirowcell{\bf probability +  expansion using aliases \\ from  \cite{hofmann}: Anchor prob. + CrossWikis + YAGO}\protect\footnotemark & 
\multirowcell{\textsf{\textbf{IBM}}, \textsf{Big\_Blue\_River\_(Kansas)}, \textsf{The\_Big\_Blue} \\ \textsf{Big\_Blue\_River\_(Indiana)},
\textsf{Big\_Blue\_(crane)}}
\\ [0.2cm] \hline

\end{tabular}}

\label{table:candgeneration}
\end{table*}
\addtocounter{footnote}{-2}


\subsubsection{Candidate Generation} \label{subsec:candgen}

An essential part of EL is candidate generation. The goal of this step is given an ambiguous entity mention, such as ``Scott Young'', to provide a list of its possible ``senses'' as specified by entities in a KG. EL is analogous to the Word Sense Disambiguation (WSD) task \cite{naviglielwsd,navigliwsd} as it also resolves lexical ambiguity. Yet in WSD, each sense of a word can be clearly defined by WordNet \cite{wordnet}, while in EL, KGs do not provide such an exact mapping between mentions and entities \cite{naviglielwsd,navigliwsd,chang-etal-2016-comparison}. Therefore, a mention potentially can be linked to any entity in a KG, resulting in a large search space, e.g. ``Big Blue'' referring to IBM. In the candidate generation step, this issue is addressed by performing effective preliminary filtering of the entity list.

Formally, given a mention $m_i$, a candidate generator provides a list of probable entities, $e_1, e_2, ..., e_k$, for each entity mention in a document.
\begin{equation}
    \mathsf{CG}: M \xrightarrow{} (e_1, e_2, ..., e_k).
\label{eq:candidategenerator}
\end{equation}

Similar to \cite{surveywei,ieeeelsurvey}, we distinguish three common candidate generation methods in neural EL: (1) based on surface form matching, (2) based on expansion with aliases, and (3) based on a prior matching probability computation. In the first approach, a candidate list is composed of entities that match various surface forms of mentions in the text 
\cite{zwicklbauer,moreno,titov1}. There are many heuristics for the generation of mention forms and matching criteria like the Levenshtein distance, n-grams, and normalization. For the example mention of ``Big Blue'', this approach would not work well, as the referent entity ``IBM'' or its long-form ``International Business Machines'' does not contain a mention string. Examples of candidate entity sets are presented in Table \ref{table:candgeneration}, where we searched a name matching of the mention ``Big Blue'' in the titles of all Wikipedia articles present in DBpedia\footnotetext{Random matches from DBpedia labels dataset -- \url{http://downloads.dbpedia.org/2016-10/core-i18n/en/labels_en.ttl.bz2}} and presented random 5 matches.
\addtocounter{footnote}{1}

In the second approach, a dictionary of additional aliases is constructed using KG metadata like disambiguation/redirect pages of Wikipedia \cite{reinforcement, zwicklbauer} or using a dictionary of aliases and/or synonyms (e.g. ``NYC'' stands for ``New York City''). This helps to improve the candidate generation recall as the surface form matching usually cannot catch such cases.  \citet{pershina} expand the given mention to the longest mention in a context found using coreference resolution. Then, an entity is selected as a candidate if its title matches the longest version of the mention, or it is present in disambiguation/redirect pages of this mention. This resource is used in many EL models, e.g. \cite{yamada,cao,griffis,elden,martins,onoe2020fine,sil}. Another well-known alternative is YAGO \cite{yago} -- an ontology automatically constructed from Wikipedia and WordNet. Among many other relations, it provides \textit{``means''} relations, and this mapping is utilized for candidate generation like in \cite{hoffart,yamada,hofmann,sil, shahbazi2018joint}. In this technique, the external information would help to disambiguate ``Big Blue'' as ``IBM''. Table \ref{table:candgeneration} shows examples of candidates generated with the help of the YAGO-means candidate mapping dataset\footnotetext{YAGO-means dataset of \citet{hoffart} -- \url{http://resources.mpi-inf.mpg.de/yago-naga/aida/download/aida_means.tsv.bz2}} used in \citet{hoffart}.
\addtocounter{footnote}{1}

The third approach to candidate generation is based on pre-calculated prior probabilities of correspondence between certain mentions and entities, $p(e|m)$. Many studies rely on mention-entity priors computed based on Wikipedia entity hyperlinks. A URL of a hyperlink to an entity page of Wikipedia determines a candidate entity, and the anchor text of the hyperlink determines a mention. Another widely-used option is CrossWikis \cite{crosswiki}, which is an extensive resource that leverages the frequency of mention-entity links in web crawl data \cite{hofmann,gupta}.

It is common to apply multiple approaches to candidate generation at once. For example, the resource constructed by \citet{hofmann} and used in many other EL methods \cite{end2end,knowbert,yamada20,shahbazi2019entity,titov2} relies on prior probabilities obtained from entity hyperlink count statistics of CrossWikis \cite{crosswiki} and Wikipedia, as well as on entity aliases obtained from the ``means'' relationship of the YAGO ontology \citet{hoffart}.
The illustrative mention ``Big Blue'' can be linked to its referent entity ``IBM'' with this method\footnotetext{We generated these examples using the source code of \citet{knowbert} -- \url{https://github.com/allenai/kb}}, as shown in Table \ref{table:candgeneration}. As another example, \citet{fang-www} utilize surface form matching and aliases. They share candidates between abbreviations and their expanded versions in the local context. The aliases are obtained from Wikipedia redirect and disambiguation pages, the Wikipedia search engine, and synonyms from WordNet \cite{wordnet}. 
Additionally, they submit mentions that are misspelled or contain multiple words to Wikipedia and Google search engines and search for the corresponding Wikipedia articles. It is also worth noting that some works also employ a candidate pruning step to reduce the number of candidates.

Recent zero-shot models \cite{zero-shot,gillick,wufacebook} perform candidate generation without external resources. Section \ref{subsubsec:domainindependent} describes them in detail.


\subsubsection{Context-mention Encoding} \label{subsec:mentionencoding}

To correctly disambiguate an entity mention, it is crucial to thoroughly capture the information from its context. The current mainstream approach is to construct a dense contextualized vector representation of a mention $\boldsymbol{y}_m$ using an encoder neural network.
\begin{equation}
    \mathsf{mENC} : (C, M)^n \xrightarrow{} (\boldsymbol{y}_{m_1}, \boldsymbol{y}_{m_2}, ...,\boldsymbol{y}_{m_n}).
\label{eq:menc}
\end{equation}

Several early techniques in neural EL utilize a convolutional encoder \cite{sun,francis2016capturing,nguyen,sorokin2018mixing},
as well as attention between candidate entity embeddings and embeddings of words surrounding a mention \cite{hofmann,titov2}. However, in recent models, two approaches prevail: recurrent networks and self-attention \cite{deep-transformer}.

A recurrent architecture with LSTM cells \cite{hochreiter1997long} that has been a backbone model for many NLP applications, is adopted to EL in \cite{end2end,martins,titov1,reinforcement,nie2018mention,gupta,sil} inter alia. \citet{gupta} concatenate outputs of two LSTM networks that independently encode left and right contexts of a mention (including the mention itself). In the same vein, \citet{sil} encode left and right local contexts via LSTMs but also pool the results across all mentions in a coreference chain and postprocess left and right representations with a tensor network. A modification of LSTM -- GRU \cite{chung2014empirical} -- is used by \citet{eshel2017named} in conjunction with an attention mechanism \cite{bahdanau2015neural} to encode left and right context of a mention. \citet{end2end} represent an entity mention as a combination of LSTM hidden states included in the mention span. \citet{titov1} simply run a bidirectional LSTM network on words complemented with embeddings of word positions relative to a target mention. \citet{shahbazi2019entity} adopt pre-trained ELMo \cite{peters2018deep} for mention encoding by averaging mention word vectors.

Encoding methods based on self-attention have recently become ubiquitous. The EL models presented in \cite{wufacebook,zero-shot,knowbert,yamada20,chen2020improving} and others rely on the outputs from pre-trained BERT layers \cite{bert} for context and mention encoding. In \citet{knowbert}, a mention representation is modeled by pooling over word pieces in a mention span. The authors also put an additional self-attention block over all mention representations that encode interactions between several entities in a sentence. Another approach to modeling mentions is to insert special tags around them and perform a reduction of the whole encoded sequence. \citet{wufacebook} reduce a sequence by keeping the representation of the special pooling symbol `[CLS]' inserted at the beginning of a sequence. \citet{zero-shot} mark positions of a mention span by summing embeddings of words within the span with a special vector and using the same reduction strategy as \citet{wufacebook}. \citet{yamada20} concatenate text with all mentions in it and jointly encode this sequence via a self-attention model based on pre-trained BERT. In addition to the simple attention-based encoder of \citet{hofmann}, \citet{chen2020improving} leverage BERT for capturing type similarity between a mention and an entity candidate. They
replace mention tokens with a special ``[MASK]'' token and extract the embedding generated for this token by BERT. A corresponding entity representation is generated by averaging multiple embeddings of mentions.



\begin{figure*}[ht]
\centering
\includegraphics[width=0.8\textwidth]{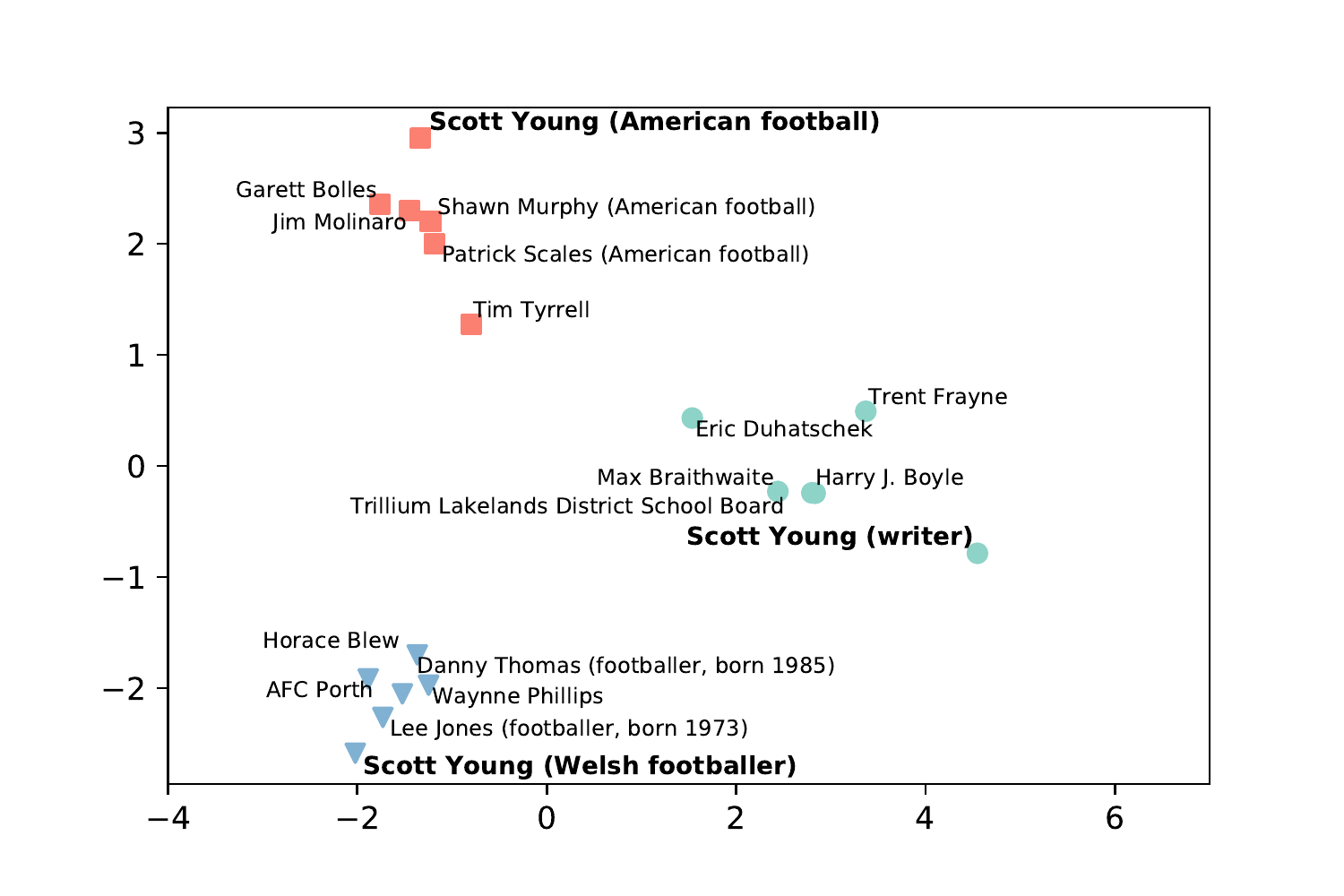}
\caption{\textbf{Visualization of entity embeddings.} Entity embedding space for entities related to the ambiguous entity mention ``Scott Young''. Three candidate entities from Wikipedia are illustrated. For each entity, their most similar $5$ entities are shown in the same colors. Entity embeddings are visualized with PCA, which is utilized to reduce dimensionality (in this example, to 2D), using pre-trained embeddings provided by \protect \citet{yamada-etal-2020-wikipedia2vec}\protect\footnotemark.}
\label{fig:pca}
\end{figure*}


\subsubsection{Entity Encoding} \label{subsec:entityrep}

To make EL systems robust, it is essential to construct distributed vector representations of entity candidates $\boldsymbol{y}_e$ in such a way that they capture semantic relatedness between entities in various aspects.
\begin{equation}
    \mathsf{eENC} : E^k \xrightarrow{} (\boldsymbol{y}_{e_1}, \boldsymbol{y}_{e_2}, ...,\boldsymbol{y}_{e_k}).
\label{eq:eenc}
\end{equation}

For instance, in Figure \ref{fig:pca}, the most similar entities for \textit{Scott Young} in the \textsf{Scott\_Young\_(American\_football)} sense are related to American football, whereas the \textsf{Scott\_Young\_(writer)} sense is in the proximity of writer-related entities. 

There are three common approaches to entity encoding in EL: (1) entity representations learned using unstructured texts and algorithms like word2vec \cite{word2vec} based on co-occurrence statistics and developed originally for embedding words; (2) entity representations constructed using relations between entities in KGs and various graph embedding methods; (3) training a full-fledged neural encoder to convert textual descriptions of entities and/or other information into embeddings.




In the first category, \citet{hofmann} collect entity-word co-occurrences statistics from two sources: entity description pages from Wikipedia; text surrounding anchors of hyperlinks to Wikipedia pages of corresponding entities. They train entity embeddings using the max-margin objective that exploits the negative sampling approach like in the word2vec model, so vectors of co-occurring words and entities lie closer to each other compared to vectors of random words and entities. Some other methods directly replace or extend mention annotations (usually anchor text of a hyperlink) with an entity identifier and straightforwardly train on the modified corpus a word representation model like word2vec \cite{zwicklbauer, doser,moreno,crossling-wikification,yamada-etal-2017-learning}. 
In \cite{moreno,hofmann,crossling-wikification,griffis}, entity embeddings are trained in such a way that entities become embedded in the same semantic space as words (or texts i.e., sentences and paragraphs \cite{yamada-etal-2017-learning}). For example, \citet{griffis} propose a distantly-supervised method that expands the word2vec objective to jointly learn words and entity representations in the shared space. The authors leverage distant supervision from terminologies that map entities to their surface forms (e.g. Wikipedia page titles and redirects or terminology from UMLS \cite{umls}).

\footnotetext{We used the English 100D embeddings from \\ \url{https://wikipedia2vec.github.io/wikipedia2vec/pretrained}}

In the second category of entity encoding methods that use relations between entities in a KG, \citet{huang} train a model that generates dense entity representations from sparse entity features (e.g. entity relations, descriptions) based on the entity relatedness. Several works expand their entity relatedness objective with functions that align words (or mentions) and entities in a unified vector space \cite{fang,yamada,yamada-etal-2020-wikipedia2vec,cao,shi,elden}, just like the methods from the first category. For example, \citet{yamada} jointly optimize three objectives to learn word and entity representations: prediction of neighbor words for the given target word, prediction of neighbor entities for the target entity based on the relationships in a KG, and prediction of neighbor words for the given entity.

Recently, knowledge graph embedding has become a prominent technique and facilitated solving various NLP and data mining tasks \cite{wang2017kgesurvey} from KG completion \cite{kgcforscholarly,transe,transh} to entity classification \cite{rescal}. For entity linking, two major graph embedding algorithms are widely adopted: DeepWalk \cite{deepwalk} and TransE \cite{transe}.

The goal of the DeepWalk \cite{deepwalk} algorithm is to produce embeddings of vertices that preserve their proximity in a graph \cite{goyal2017graph}. It first generates several random walks for each vertex in a graph. The generated walks are used as training data for the skip-gram algorithm. Like in word2vec for language modeling, given a vertex, the algorithm maximizes the probabilities of its neighbors in the generated walks. \citet{fastandaccurate,sevgili} leverage DeepWalk-based graph embeddings built from DBpedia \cite{dbpedia} for entity linking. \citet{fastandaccurate} use entity embeddings to compute cosine similarity scores of candidate entities in global entity linking. \citet{sevgili} show that combining graph and text-based embeddings can slightly improve the performance of neural entity disambiguation when compared to using only text-based embeddings.

The goal of the TransE \cite{transe} algorithm is to construct embeddings of both vertices and relations in such a way that they are compatible with the facts in a KG \cite{wang2017kgesurvey}. Consider the facts in a KG are represented in the form of triples (i.e. head entity, relation, tail entity). If a fact is contained in a KG, the TransE margin-based ranking criterion facilitates the presence of the following correspondence between embeddings: $head+relation \approx tail$. This means that the relationship in a KG should be a linear translation in the embedding space of entities. At the same time, if there is no such fact in a KG, this functional relationship should not hold. 
The TransE-based entity representations constructed from Wikidata \cite{wikidata} and Freebase \cite{freebase} have been used for entity representation in language modeling \cite{ernie} and in several works on EL \cite{debayan,sorokin2018mixing,end2endkge}. 
\citet{debayan,sorokin2018mixing} utilize Wikidata-based entity embeddings as an input component of neural models along with other types of information about entities. The ablation study conducted by \citet{debayan} show that the TransE entity embeddings are the most important features for their entity linking model. They attribute this finding to the fact that graph embeddings contain rich information about the KG structure. Similarly, \citet{sorokin2018mixing} find that without KG structure information, their entity linker experiences a big performance drop. \citet{end2endkge} integrate knowledge graph embeddings built from Freebase and word embeddings in a single end-to-end model that solves entity and relation linking tasks jointly. The quantitative analysis shows that their KG-embedding-based method helps to pick correct entity candidates. Recently, \citet{wu-dgcn} also utilize TransE embeddings with other types of entity embeddings, like \citet{hofmann} or dynamic representation, to compute pairwise entity relatedness scores. 


There are many other techniques for KG embedding: \cite{node2vec,transh,rescal,trouillon2016complex,yang2015embedding,convolutionalkge} inter alia and very recent 5*E \cite{5star}, which is designed to preserve complex graph structures in the embedding space. However, they are not widely used in entity linking right now. A detailed overview of all graph embedding algorithms is out of the scope of the current work. We refer the reader to the previous surveys on this topic \cite{goyal2017graph, graphembedssurvey,wang2017kgesurvey,Ruffinelli2020You} and consider integration of novel KG embedding techniques in EL models a promising research direction.

In the last category, we place methods that produce entity representations using other types of information like entity descriptions and entity types. Often, an entity encoder is a full-fledged neural network, which is a part of an entity linking architecture. \citet{sun} use a neural tensor network to encode interactions between surface forms of entities and their category information from a KG. In the same vein, \citet{francis2016capturing} and \citet{nguyen} construct
entity representations by encoding titles and entity description pages with convolutional neural networks. In addition to a convolutional encoder for entity descriptions, \citet{gupta} also include an encoder for fine-grained entity types by using the type set of FIGER \cite{figer}.
\citet{gillick} construct entity representations by encoding entity page titles, short entity descriptions, and entity category information with feed-forward networks. \citet{titov1} use only entity type information from a KG and a simple feed-forward network for entity encoding.
\citet{hou-etal-2020-improving} also leverage entity types. However, instead of relying on existing type sets like in \cite{gupta}, they construct custom fine-grained semantic types using words from starting sentences of Wikipedia pages. To represent entities, they first average the word vectors of entity types and then linearly aggregate them with embeddings of \citet{hofmann}.

Recent works leverage deep language models like BERT \cite{bert} or ELMo \cite{peters2018deep} for encoding entities. \citet{nie2018mention} use an architecture based on a recurrent network for obtaining entity representations from Wikipedia entity description pages. Subsequently, several models adopt BERT for the same purpose  \cite{zero-shot,wufacebook} inter alia.
\citet{yamada20} propose a masked entity prediction task, where a model based on the BERT architecture learns to predict randomly masked input entities. 
This task makes the model learn also how to generate entity representations along with standard word representations. 
\citet{shahbazi2019entity} introduce E-ELMo that extends the ELMo model \cite{peters2018deep} with an additional objective. The model is trained in a multi-task fashion: to predict next/previous words, as in a standard bidirectional language model, and to predict the target entity when encountering its mentions. As a result, besides the model for mention encoding, entity representations are obtained. \citet{mulang2020evaluating} use bidirectional Transformers to jointly encode context of a mention, a candidate entity name, and multiple relationships of a candidate entity from a KG verbalized into textual triples: ``[subject] [predicate] [object]''. The input sequence of the encoder is composed simply by appending all these types of information delimited by a special separator token. 

%

\begin{figure*}
    \centering
    \includegraphics[width=.65\textwidth]{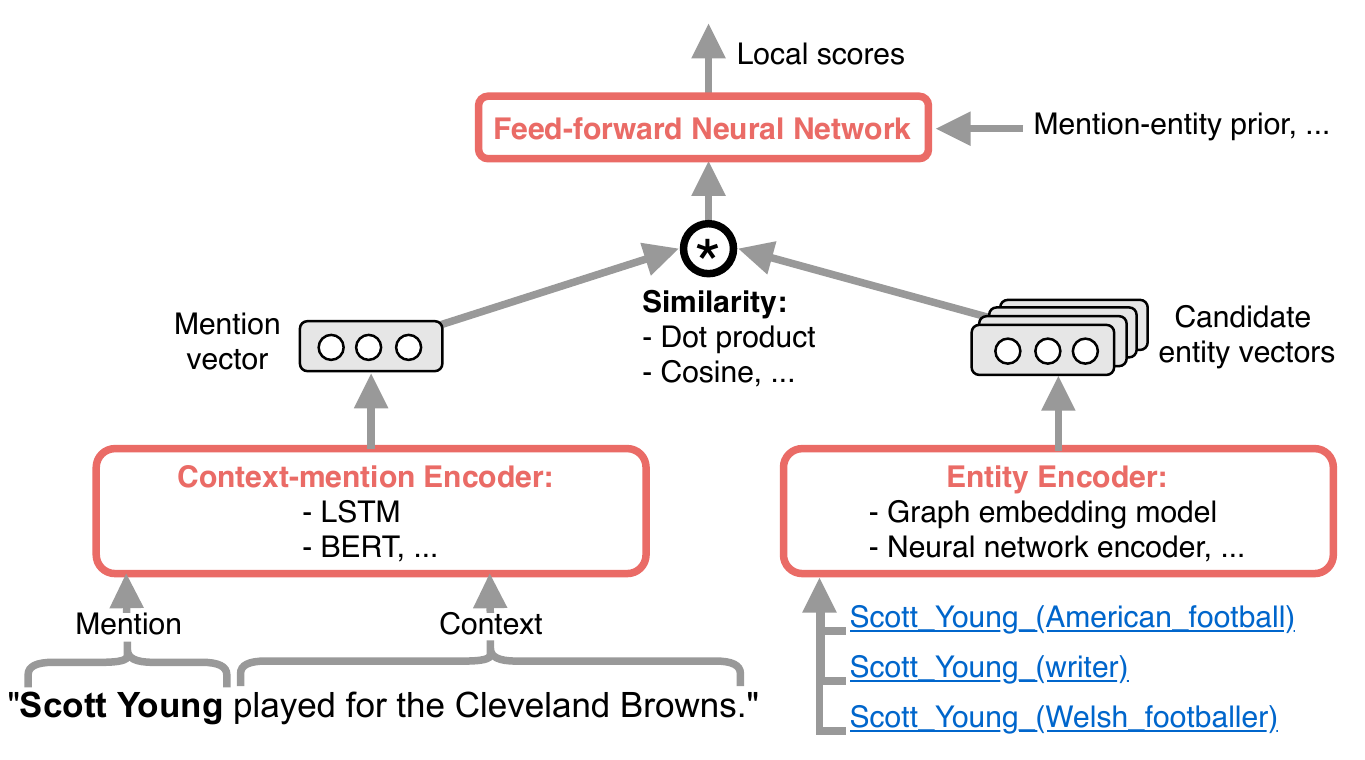}
    \caption{\textbf{Entity ranking}. A generalized entity candidate ranking neural architecture: entity candidates are ranked according their appropriateness for a particular mention in the current context.
    }
    \label{fig:ranking}
\end{figure*}

\subsubsection{Entity Ranking} \label{subsec:entitydisambiguation}

The goal of this stage is given a list of entity candidates $(e_1, e_2, ..., e_k)$ from a KG and a context $C$ with a mention $M$ to rank these entities assigning a score to each of them, as in Equation \ref{eq:entityranking}, where $n$ is a number of entity mentions in a document, $k$ is a number of candidate entities. Figure \ref{fig:ranking} depicts the typical architecture of the ranking component.
\begin{equation}
    \mathsf{RNK} : ((e_1, e_2, ..., e_k), C, M)^n \xrightarrow{} \mathbb{R}^{n \times k}.
\label{eq:entityranking}
\end{equation}

The mention representation $ \boldsymbol{y}_m$ generated in the mention encoding step is compared with candidate entity representations $ \boldsymbol{y}_{e_i} (i=1,2,\dots,k) $ according to the similarity measure $s(m, e_i)$. Entity representations can be pre-trained (see Section \ref{subsec:entityrep}) or generated by another encoder as in some zero-shot approaches (see Section \ref{subsubsec:domainindependent}). The BERT-based model of \citet{yamada20} simultaneously learns how to encode mentions and entity embeddings in the unified architecture. 

Most of the state-of-the-art studies compute similarity $s(m,e)$ between representations of a mention $m$ and an entity $e$  using a dot product as in \cite{hofmann,gupta,end2end,knowbert,wufacebook}:
\begin{equation}
s\left(m, e_{i}\right)=\boldsymbol{y}_{m} \cdot \boldsymbol{y}_{e_{i}};
\label{eq:dot}
\end{equation}
or cosine similarity as in \cite{sun,francis2016capturing,gillick}: 
\begin{equation}
s\left(m, e_{i}\right)=\cos(\boldsymbol{y}_{m}, \boldsymbol{y}_{e_{i}})=\frac{ \boldsymbol{y}_{m} \cdot \boldsymbol{y}_{e_{i}} }{ \|\boldsymbol{y}_{m}\| \cdot \|\boldsymbol{y}_{e_{i}}\|}. 
\label{eq:cosine}
\end{equation}

The final disambiguation decision is inferred via a probability distribution $P(e_i|m)$, which is usually approximated by a softmax function over the candidates. The calculated similarity score or probability can be combined with mention-entity priors obtained during the candidate generation phase \cite{francis2016capturing,hofmann,end2end} or other features $f(e_i,m)$ such as  various similarities, a string matching indicator, and entity types or type similarity \cite{francis2016capturing,shahbazi2018joint,sil,shahbazi2019entity,yang2019,chen2020improving}. One of the common techniques for that is to use an additional one or two-layer feedforward network $\phi(\cdot,\cdot)$ \cite{francis2016capturing,hofmann,shahbazi2019entity}. The obtained local similarity score $\Phi(e_i,m)$ or the probability distribution can be further utilized for global scoring (see Section \ref{subsubsec:localglobal}). 
\begin{equation}
    P(e_i|m) = \frac{\exp(s(m,e_i))}{\sum_{i=1}^k{\exp(s(m, e_i))}}.
\end{equation}
\vspace{-0.5cm}
\begin{equation}
    \Phi(e_i,m) = \phi(P(e_i|m), f(e_i, m)).
\end{equation}

There are several approaches to framing a training objective in the literature on EL. Consider that we have $k$ candidates for the target mention $m$, one of which is a true entity $e_*$. In some works, the models are trained with the standard negative log-likelihood objective like in classification tasks \cite{zero-shot,wufacebook}. However, instead of classes, negative candidates are used:
\begin{equation}
\mathcal{L}\left(m\right)=-s\left(m, e_*\right)+\log \sum_{i=1}^{k} \exp \left(s\left(m, e_{i}\right)\right).
\label{eq:loglikelihood}
\end{equation}

Instead of the the negative log-likelihood, some works use variants of a ranking loss. The idea behind such an approach is to enforce a positive margin $\gamma > 0$ between similarity scores of mentions to positive and negative candidates \cite{hofmann,end2end,knowbert}:
\begin{equation}
\mathcal{L}(m) = \sum_i{\ell(e_i,m)}, \text{ where }
\label{eq:rankingloss}
\end{equation}
\vspace{-0.3cm}
\begin{equation}
\ell(e_i, m)=\left[\gamma-\Phi\left(e_*, m\right)+\Phi(e_i, m)\right]_{+}.
\label{eq:rankinglossexpand}
\end{equation}
or
\begin{equation}
\begin{array}{l}
\ell(e_i, m)= \\
\quad\left\{\begin{array}{ll}
\left[ \gamma-\Phi(e_i, m) \right]_{+}, & \text { if } e_i \text{ equal } e_*  \\
\left[ \Phi(e_i, m) \right]_{+}, & \text { otherwise. }
\end{array}\right.
\end{array}
\label{eq:rankinglossor}
\end{equation}

\begin{figure*}
    \centering
    \begin{tikzpicture}[
  level 1/.style={sibling distance=41mm},
  edge from parent/.style={->,draw, mygrey, thick, },
  >=latex]

\node[root, text width=4cm] {\footnotesize \textbf{3.2 - Modifications of the General Architecture}}
  child {node[level 2] (c1) {\footnotesize \textbf{3.2.1 - Joint Entity Mention Detection and Disambiguation Architecture}}}
  child {node[level 2] (c2) {\footnotesize \textbf{3.2.2 - \\ Global Context Architecture}}}
  child {node[level 2] (c3) {\footnotesize \textbf{3.2.3 - Domain Independent Architecture}}}
  child {node[level 2] (c4) {\footnotesize \textbf{3.2.4 - \\ Cross-lingual Architecture}}};

\begin{scope}[every node/.style={level 3}]
\node [below of = c1, xshift=15pt, text width=3.5cm] (c11) {\footnotesize candidate based \cite{end2end,knowbert}};
\node [below of = c11, xshift=-10pt, yshift=10pt] (c12) {\footnotesize  multitask learning \cite{martins}};
\node [below of = c12, yshift=10pt] (c13) {\footnotesize  sequence labeling \cite{broscheit2020investigating}};

\node [below of = c2, xshift=8pt,  text width=3.1cm, yshift=8pt] (c21) {\footnotesize graph based \cite{zwicklbauer,cao2018}};
\node [below of = c21, xshift=6pt, text width=3.5cm, yshift=8pt] (c22) {\footnotesize maximization of CRF potentials \cite{hofmann,le2018}};
\node [below of = c22, xshift=22pt, text width=5cm,yshift=9pt] (c23) {\footnotesize sequential decision task \cite{reinforcement,yang2019,yamada20}};
\node [below of = c23, xshift=-21pt, text width=3.5cm,yshift=15pt] (c24) {\footnotesize others  \cite{end2end,fang}};

\node [below of = c3,  xshift=6pt, yshift=5pt, text width=3cm] (c31) {\footnotesize distant learning \cite{titov1,titov2}};
\node [below of = c31,xshift=-3pt, yshift=10pt] (c32) {\footnotesize zero-shot \cite{zero-shot, gillick,wufacebook}};

\node [below of = c4, xshift=4pt] (c41) {\footnotesize representation based \cite{pan2017cross, crossling-wikification}};
\node [below of = c41, yshift=5pt] (c42) {\footnotesize zero-shot \cite{upadhyayjoint,sil}};
\end{scope}

\foreach \value in {1,2,3}
  \draw[->, mygrey,thick] (c1.202) |- (c1\value.west);

\foreach \value in {1,2,3,4}
  \draw[->, mygrey,thick] (c2.194) |- (c2\value.west);

\foreach \value in {1,2}
  \draw[->, mygrey,thick] (c3.195) |- (c3\value.west);
\foreach \value in {1,2}
  \draw[->, mygrey,thick] (c4.195) |- (c4\value.west);
\end{tikzpicture}
    \caption{\textbf{Reference map of the modifications of the general architecture for neural EL.} The categorization of each modification with various design choices and example references illustrating each choice. Sections \ref{subsubsec:domainindependent} and \ref{subsubsec:cross-lingual} are categorized based on their EL solutions, here.}
    \label{fig:taxonomymodificationstikz}
\end{figure*}
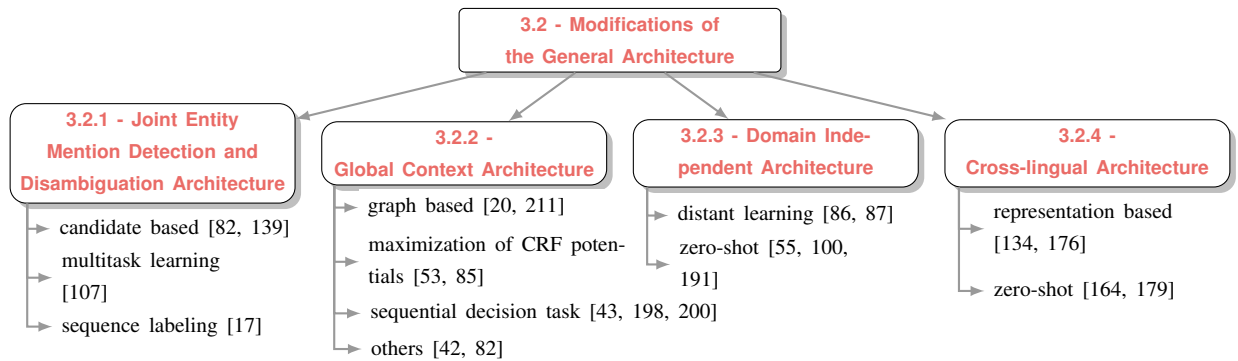

\subsubsection{Unlinkable Mention Prediction} \label{subsec:nil}

The referent entities of some mentions can be absent in the KGs, e.g. there is no Wikipedia entry about \textit{Scott Young} as a cricket player of the Stenhousemuir cricket club.\footnote{Information about \textit{Scott Young} as a cricket player: \url{https://www.stenhousemuircricketclub.com/teams/171906/player/scott-young-1828009}} Therefore, an EL system should be able to predict the absence of a reference if a mention appears in specific contexts, which is known as the NIL prediction task: 
\begin{equation}
    \mathsf{NILp} : (C, M)^n \xrightarrow{} \{0,1\}^{n}.
\label{eq:nil}
\end{equation}

The NIL prediction task is essentially a classification with a reject option~\cite{hellman1970nearest,fumera2000reject,herbei2006classification}.
There are four common ways to perform NIL prediction. Sometimes a candidate generator does not yield any corresponding entities for a mention; such mentions are trivially considered unlikable \cite{crossling-wikification,sil}. One can set a threshold for the best linking probability (or a score), below which a mention is considered unlinkable \cite{knowbert,lazic}. Some models introduce an additional special ``NIL'' entity in the ranking phase, so models can predict it as the best match for the mention \cite{end2end}. It is also possible to train an additional binary classifier that accepts mention-entity pairs after the ranking phase, as well as several additional features (best linking score, whether mentions are also detected by a dedicated NER system, etc.), as input and makes the final decision about whether a mention is linkable or not \cite{moreno,martins}.

\subsection{Modifications of the General Architecture} \label{subsec:modifications}

This section presents the most notable modifications and improvements of the general architecture of neural entity linking models presented in Section~\ref{subsec:generalapproaches} and Figures~\ref{fig:edpipeline} and~\ref{fig:ranking}. The categorization of each modification is summarized in Figure \ref{fig:taxonomymodificationstikz}.

\subsubsection{Joint Entity Mention Detection and Disambiguation} \label{subsubsec:jointedandel}

While it is common to separate the mention detection (cf. Equation~\ref{eq:recognition}) and entity disambiguation stages (cf. Equation~\ref{eq:disambiguator}), as illustrated in Figure~\ref{fig:systemdesc}, a few systems provide \textit{joint} solutions for entity linking where entity mention detection and disambiguation are done at the same time by the same model. Formally, the task becomes to detect a mention $m_i \in M$ and predict an entity $e_i \in E$ for a given context $c_i \in C$, for all $n$ entity mentions in the context: 
\begin{equation}
    \mathsf{EL} : C \xrightarrow{} (M, E)^n.
\label{eq:jointered}
\end{equation}

Undoubtedly, solving these two problems simultaneously makes the task more challenging. However, the interaction between these steps can be beneficial for improving the quality of the overall pipeline due to their natural mutual dependency. While first competitive models that provide joint solutions were probabilistic graphical models \cite{luo2015joint,nguyen2016j}, we focus on purely neural approaches proposed recently \cite{sorokin2018mixing,end2end,knowbert,martins,broscheit2020investigating,chen-etal-2020-contextualized,poerner-etal-2020-e,de2020autoregressive}.

The main difference of joint models is the necessity to produce also mention candidates. For this purpose, \citet{end2end} and \citet{knowbert} enumerate all spans in a sentence with a certain maximum width, filter them by several heuristics (remove mentions with stop words, punctuation, ellipses, quotes, and currencies), and try to match them to a pre-built index of entities used for the candidate generation. If a mention candidate has at least one corresponding entity candidate, it is further treated by a ranking neural network that can also discard it by considering it unlinkable to any entity in a KG (see Section \ref{subsec:entitydisambiguation}). Therefore, the decision during the entity disambiguation phase affects mention detection. In a similar fashion, \citet{sorokin2018mixing} treat each token n-gram up to a certain length as a possible mention candidate. They use an additional binary classifier for filtering candidate spans, which is trained jointly with an entity linker. \citet{debayan} also enumerates all possible n-grams and expands each of them with candidate entities, which results in a long sequence of points corresponding to a candidate entity for a particular mention n-gram. This sequence is further processed by a single-layer BiLSTM pointer network \cite{pointernetworks} that generates index numbers of potential entities in the input sequence. \citet{li-etal-2020-efficient} consider various possible spans as mention candidates and introduce a loss component for boundary detection, which is optimized along with the loss for disambiguation. 

\citet{martins} describe the approach with tighter integration between detection and linking phases via multi-task learning. The authors propose a stack-based bidirectional LSTM network with a shift-reduce mechanism and attention for entity recognition that propagates its internal states to the linker network for candidate entity ranking. The linker is supplemented with a NIL predictor network. The networks are trained jointly by optimizing the sum of losses from all three components. 


\begin{figure*}[t]
\centering
\includegraphics[width=1.0\textwidth]{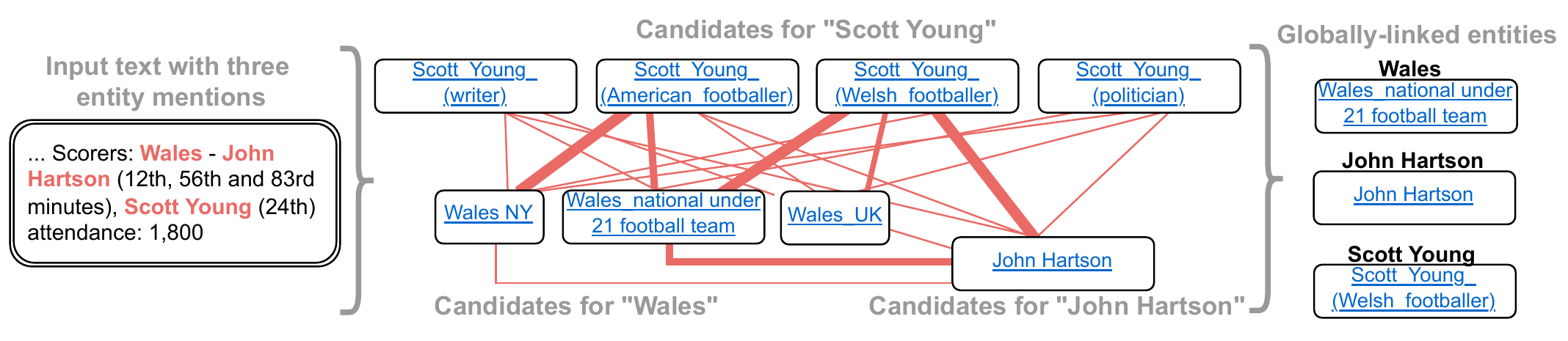}
\caption{\textbf{Global entity disambiguation}. The global entity linking resolves all mentions simultaneously based on entity coherence. Bolder lines indicate expected higher degrees of entity-entity similarity.
}
\label{fig:globalel}

\end{figure*}


\citet{broscheit2020investigating} goes further by suggesting a completely end-to-end method that deals with mention detection and linking jointly without explicitly executing a candidate generation step. In this work, the EL task is formulated as a sequence labeling problem, where each token in the text is assigned an entity link or a NIL class. They leverage a sequence tagger based on pre-trained BERT for this purpose. This simplistic approach does not supersede \cite{end2end} but outperforms the baseline, in which candidate generation, mention detection, and linking are performed independently. In the same vein,  \citet{chen-etal-2020-contextualized} use a sequence tagging framework for joint entity mention detection and disambiguation. However, they experiment with both settings: when a candidate list is available and not, and demonstrate that it is possible to achieve high linking performance without candidate sets. Similar to \citet{li-etal-2020-efficient}, they optimize the joint loss for linking and mention boundary detection.

\citet{poerner-etal-2020-e} propose a model E-BERT-MLM, in which they repurpose the masked language model (MLM) objective for the selection of entity candidates in an end-to-end EL pipeline. The candidate mention spans and candidate entity sets are generated in the same way as in \cite{end2end}. For candidate selection, E-BERT-MLM inserts a special ``[E-MASK]'' token into the text before the considered candidate mention span and tries to restore an entity representation for it. The model is trained by minimizing the cross-entropy between the generated entity distribution of the potential spans and gold entities. In addition to the standard BERT architecture, the model contains a linear transformation pre-trained to align entity embeddings with embeddings of word-piece tokens.

\citet{de2020autoregressive} recently have proposed a generative approach to performing mention detection and disambiguation jointly. Their model, which is based on BART \cite{lewis-etal-2020-bart}, performs a sequence-to-sequence autoregressive generation of text markup with information about mention spans and links to entities in a KG. The generation process is constrained by a markup format and a candidate set, which is retrieved from standard pre-built candidate resources. Most of the time, the network works in a copy-paste regime when it copies input tokens into the output. When it finds a beginning of a mention, the model marks it with a square bracket, copies all tokens of a mention, adds a finishing square bracket, and generates a link to an entity. Although this approach to EL, at the first glance, is counterintuitive and completely different from the solutions with a standard bi-encoder architecture, this model achieves near state-of-the-art results for joint MD and ED and competitive performances on ED-only benchmarks. However, as it is shown in the paper, to achieve such impressive results, the model had to be pre-trained on a large annotated Wikipedia-based dataset \cite{wufacebook}. The authors also note that the memory footprint of the proposed model is much smaller than that of models based on the standard architecture due to no need for storing entity embeddings.


\subsubsection{Global Context Architectures} \label{subsubsec:localglobal}

Two kinds of contextual information are available in entity disambiguation: local and global. In local approaches to ED, each mention is disambiguated independently based on the surrounding words, as in the following function: 
\begin{equation}
\mathsf{LED}: (M, C) \xrightarrow{} E.
\label{eq:local}
\end{equation}

Global approaches to ED take into account semantic consistency (coherence) across multiple entities in a context. In this case, all $q$ entity mentions in a group are disambiguated interdependently: a disambiguation decision for one entity is affected by decisions made for other entities in a context as illustrated in Figure \ref{fig:globalel} and Equation \ref{eq:global}. 
%
\begin{equation}
    \mathsf{GED} : ((m_1, m_2, ..., m_q), C)\xrightarrow{} E^q.
\label{eq:global}
\end{equation}

In the example presented in Figure \ref{fig:globalel}, the consistency score between correct entity candidates: the \textit{national football team} sense of \textit{Wales} and the \textit{Welsh footballer} sense of \textit{Scott Young} and \textit{John Hartson}, is expected to be higher than between incorrect ones. 

Besides involving consistency, the considered context of a mention in global methods is usually larger than in local ones or even extends to the whole document. Although modeling consistency between entities and the extra information of the global context improves the disambiguation accuracy, the number of possible entity assignments is combinatorial \cite{pboh}, which results in high time complexity of disambiguation \cite{yang2019,hofmann}. Another difficulty is an attempt to assign an entity its consistency score since this score is not possible to compute in advance due to the simultaneous disambiguation \cite{yamada}.

The typical approach to global disambiguation is to generate a graph including candidate entities of mentions in a context and perform some graph algorithms, like random walk algorithms (e.g. PageRank \cite{pagerank}) or graph neural networks, over it to select highly consistent entities \cite{zwicklbauer, doser,pershina,guo18}. Recently, \citet{ijcai2019-740} propose a neural recurrent random walk network learning algorithm based on the transition matrix of candidate entities containing relevance scores, which are created from hyperlinks information and cosine similarity of entities. \citet{cao2018} construct a subgraph from the candidates of neighbor mentions, integrate local and global features of each candidate, and apply a graph convolutional network over this subgraph. In this approach, the graph is static, which would be problematic in such cases that two mentions would co-occur in different documents with different topics, however, the produced graphs will be the same, and so, could not catch the different information \cite{wu-dgcn}. To address it, \citet{wu-dgcn} propose a dynamic graph convolution architecture, where entity relatedness scores are computed and updated in each layer based on the previous layer information (initialized with some features, including context scores) and entity similarity scores. \citet{globerson} introduce a model with an attention mechanism that takes into account only the subgraph of the target mention, rather than all interactions of all the mentions in a document and restrict the number of mentions with an attention.

Some works approach global ED by maximizing the Conditional Random Field (CRF) potentials, where the first component $\Psi$ represents a local entity-mention score, and the other component $\Phi$ measures coherence among selected candidates \cite{hofmann,pboh,le2018,titov2}, as defined in \citet{hofmann}:
\begin{equation}
    g(e,m,c) = \sum_{i=1}^{n}\Psi(e_i, m_i, c_i) + \sum_{i<j}\Phi(e_i, e_j).
\label{eq:crf}
\end{equation}

However, model training and its exact inference are NP-hard. \citet{hofmann} utilize truncated fitting of loopy belief propagation \cite{pboh,globerson} with differentiable and trainable message passing iterations using pairwise entity scores to reduce the complexity. \citet{le2018} expand it in a way that pairwise scores take into account relations of mentions (e.g. \textsf{located\_in}, or coreference: the mentions are coreferent if they refer to the same entity) by modeling relations between mentions as latent variables. \citet{shahbazi2018joint} develop a greedy beam search strategy, 
which starts from a locally optimal initial solution and is improved by searching for possible corrections with the focus on the least confident mentions. 

Despite the optimizations proposed like in some aforementioned works, taking into account coherence scores among candidates of all mentions at once can be prohibitively slow. It also can be malicious due to erroneous coherence among wrong entities \cite{reinforcement}. For example, if two mentions have coherent erroneous candidates, this noisy information may mislead the final global scoring. 
To resolve this issue, some studies define the global ED problem as a sequential decision task, where the disambiguation of new entities is based on the already disambiguated ones with high confidence. \citet{reinforcement} train a policy network for sequential selection of entities using reinforcement learning. The disambiguation of mentions is ordered according to the local score, so the mentions with high confident entities are resolved earlier. The policy network takes advantage of output from the LSTM global encoder that maintains the information about earlier disambiguation decisions. \citet{yang2019} also utilize reinforcement learning 
for mention disambiguation. They use an attention model to leverage knowledge from previously linked entities. The model dynamically selects the most relevant entities for the target mention and calculates the coherence scores. \citet{yamada20} iteratively predict entities for yet unresolved mentions with a BERT model, while attending on the previous most confident entity choices. Similarly, \citet{mrc} sort mentions based on their ambiguity degrees produced by their BERT-based local model and update query/context based on the linked entities so that the next prediction can leverage the previous knowledge. They also utilize a gate mechanism to control historical cues -- representations of linked entities. \citet{yamada} and \citet{elden} measure the similarity first based on unambiguous mentions and then predict entities for complex cases. \citet{nguyen} use an RNN to implicitly store information about previously seen mentions and corresponding entities. They leverage the hidden states of the RNN to reach this information as a feature for the computation of the global score. 
\citet{crossling-wikification} directly use embeddings of previously linked entities as features for the disambiguation model. Recently, \citet{fang-www} combine sequential approaches with graph based methods, where the model dynamically changes the graph depending on the current state. The graph is constructed with previously resolved entities, current candidate entities, and subsequent mention's candidates. The authors use a graph attention network over this graph to make a global scoring. As explained before, \citet{wu-dgcn} also 
change the entity graph dynamically depending on the outputs from previous layers of a GCN. \citet{zwicklbauer} include to the candidates graph a topic node created from the set of already disambiguated entities.

Some studies, for example, 
\citet{end2end} model the coherence component as an additional feed-forward neural network that uses the similarity score between the target entity and an average embedding of the candidates with a high local score. \citet{fang} use the similarity score between the target entity and its surrounding entity candidates in a specified window as a feature for the disambiguation model. 

Another approach that can be considered as global is to make use of a document-wide context, which usually contains more than one mention and helps to capture the coherence implicitly instead of explicitly designing an entity coherence component \cite{knowbert,gupta,moreno,francis2016capturing}.

\subsubsection{Domain-Independent Architectures} \label{subsubsec:domainindependent}

Domain independence is one of the most desired properties of EL systems. Annotated resources are very limited and exist only for a few domains. Obtaining labeled data in a new domain requires much labor. Earlier, this problem is tackled by few domain-independent approaches based on unsupervised \cite{wang2015,cao,griffis} and semi-supervised models \cite{lazic}. Recent studies provide solutions based on distant learning and zero-shot methods.

\citet{titov2,titov1} propose distant learning techniques that use only unlabeled documents. They rely on the weak supervision coming from a surface matching heuristic, and the EL task is framed as binary multi-instance learning. The model learns to distinguish between a set of positive entities and a set of random negatives. The positive set is obtained by retrieving entities with a high word overlap with the mention and that have relations in a KG to candidates of other mentions in the sentence. While showing promising performance, which in some cases rivals results of fully supervised systems, these approaches require either a KG describing relations of entities \cite{titov1} or mention-entity priors computed from entity hyperlink statistics extracted from Wikipedia \cite{titov2}.  

Recently proposed zero-shot techniques \cite{zero-shot,wufacebook,yao-etal-2020-zero,tang2021bidirectional} tackle problems related to adapting EL systems to new domains. In the zero-shot setting, the only entity information available is its description. As well as in other settings, texts with mention-entity pairs are also available. The key idea of zero-shot methods is to train an EL system on a domain with rich labeled data resources and apply it to a new domain with only minimal available data like descriptions of domain-specific entities. One of the first studies that proposes such a technique is \citet{gupta} (not purely zero-shot because they also use entity typings). Existing zero-shot systems do not require such information resources as surface form dictionaries, prior entity-mention probabilities, KG entity relations, and entity typing, which makes them particularly suited for building domain-independent solutions. However, the limitation of information sources raises several challenges. 

Since only textual descriptions of entities are available for the target domain, one cannot rely on pre-built dictionaries for candidate generation. All zero-shot works rely on the same strategy to tackle candidate generation: pre-compute representations of entity descriptions (sometimes referred to as caching), compute a representation of a mention, and calculate its similarity with all the description representations. Pre-computed representations of descriptions save a lot of time at the inference stage. Particularly, \citet{zero-shot} use the BM25 information retrieval formula \cite{jones2000probabilistic}, which is a similarity function for count-based representations. 

A natural extension of count-based approaches is embeddings. The method proposed by \citet{gillick}, which is a predecessor of zero-shot approaches, uses average unigram and bigram embeddings followed by dense layers to obtain representations of mentions and descriptions. The only aspect that separates this approach from pure zero-shot techniques is the usage of entity categories along with descriptions to build entity representations. Cosine similarity is used for the comparison of representations. Due to the computational simplicity of this approach, it can be used in a single stage fashion where candidate generation and ranking are identical. For further speedup, it is possible to make this algorithm two-staged. In the first stage, an approximate search can be used for candidate set retrieval. In the second stage, the retrieved smaller set can be used for exact similarity computation. Instead of simple embeddings, \citet{wufacebook} suggest using a BERT-based bi-encoder for candidate generation. Two separate encoders generate representations of mentions and entity descriptions. Similar to the previous work, the candidate selection is based on the score obtained via a dot-product of mention/entity representations.

For entity ranking, a very simple embedding-based approach of \citet{gillick} described above shows very competitive scores on the TAC KBP-2010 benchmark, outperforming some complex neural architectures. The recent studies of \citet{zero-shot} and \citet{wufacebook} utilize a BERT-based cross-encoder to perform joint encoding of mentions and entities. The cross-encoder takes a concatenation of a context with a mention and an entity description to produce a scalar score for each candidate. The cross-attention helps to leverage the semantic information from the context and the definition on each layer of the encoder network \cite{humeau2019poly,reimers-gurevych-2019-sentence}. In both studies, cross-encoders achieve superior results compared to bi-encoders and count-based approaches. For entity linking, cross-attention between mention context representations and entity descriptions is also used by \citet{nie2018mention}. However, they leverage recurrent architectures for encoding. \citet{yao-etal-2020-zero} introduce a small tweak of positional embeddings in the \citet{zero-shot}'s architecture aimed at better handling long contexts. \citet{tang2021bidirectional} address the problem of the limited size of the mention context and the entity description that could be processed by the standard BERT model. They argue that the input size of 512 tokens is not enough to capture context and entity description relatedness since the evidence for linking could scatter in different paragraphs and suggest a novel architecture that resolves this problem. Roughly speaking, their model splits the context of a mention and entity description into multiple paragraphs, performs cross-attention between representations of these paragraphs, and aggregates the results for disambiguation. The experimental results show that their model substantially improves the zero-shot performance keeping the inference time in an acceptable range.

Evaluation of zero-shot systems requires data from different domains. \citet{zero-shot} proposes the \textit{Zero-shot EL}\footnote{\url{https://github.com/lajanugen/zeshel}} dataset, constructed from several Wikias\footnote{\url{https://www.wikia.com}}. In the proposed setting, training is performed on one set of Wikias while evaluation is performed on others. \citet{gillick} construct the Wikinews dataset. This dataset can be used for evaluation after training on Wikipedia data.

Clearly, heavy neural architectures pre-trained on general-purpose open corpora substantially advance the performance of zero-shot techniques. As highlighted by \citet{zero-shot} further unsupervised pre-training on source data, as well as on the target data is beneficial. The development of better approaches to the utilization of unlabeled data might be a fruitful research direction. Furthermore, closing the performance gap of entity ranking between a fast representation based bi-encoder and a computationally intensive cross-encoder is an open question.

\subsubsection{Cross-lingual Architectures} \label{subsubsec:cross-lingual}

An abundance of labeled data for EL in English contrasts with the amount of data available in other languages. The cross-lingual EL (sometimes called XEL) methods \cite{crosslingoverview} aim at overcoming the lack of annotation for resource-poor languages by leveraging supervision coming from their resource-rich counterparts. Many of these methods are feasible due to the presence of a unique source of supervision for EL -- Wikipedia, which is available for a variety of languages. The inter-language links in Wikipedia that map pages in one language to equivalent pages in another language also help to map corresponding entities in different languages.







Challenges in XEL start at candidate generation and mention detection steps since a resource-poor language can lack mappings between mention strings and entities. In addition to the standard mention-entity priors based on inter-language links \cite{crossling-wikification,sil,upadhyayjoint}, candidate generation can be approached by mining a translation dictionary \cite{pan2017cross}, training a translation and alignment model \cite{tsai2018learning,upadhyay-etal-2018-bootstrapping}, or applying a neural character-level string matching model \cite{rijhwani2019zero,zhou-etal-2019-towards}. In the latter approach, the model is trained to match strings from a high-resource pivot language to strings in English. If a high-resource pivot language is similar to the target low-resource one, such a model is able to produce reasonable candidates for the latter. The neural string matching approach can be further improved with simpler average n-gram encoding and extending entity-entity pairs with mention-entity examples \cite{zhou2020improving}. 
Such an approach can also be applied to entity recognition \cite{cotterell-duh-2017-low}. \citet{fu2020design} criticize methods that solely rely on Wikipedia due to the lack of inter-language links for resource-poor languages. They propose a candidate generation method that leverages results from querying online search engines (Google and Google Maps) and show that due to its much higher recall compared to other methods, it is possible to substantially increase the performance of XEL.

There are several approaches to candidate ranking that take advantage of cross-lingual data for dealing with the lack of annotated examples. \citet{pan2017cross} use the Abstract Meaning Representation (AMR) \cite{banarescu2013abstract} statistics in English Wikipedia and mention context for ranking. To train an AMR tagger, pseudo-labeling \cite{lee2013pseudo} is used. \citet{crossling-wikification} train monolingual embeddings for words and entities jointly by replacing every entity mention with corresponding entity tokens. Using the inter-language links, they learn the projection functions from multiple languages into the English embedding space. For ranking, context embeddings are averaged, projected into the English space, and compared with entity embeddings. The authors demonstrate that this approach helps to build better entity representations and boosts the EL accuracy in the cross-lingual setting by more than 1\% for Spanish and Chinese. \citet{sil} propose a method for zero-shot transfer from a high-resource language. The authors extend the previous approach with the least squares objective for embedding projection learning, the CNN context encoder, and a trainable re-weighting of each dimension of context and entity representations. The proposed approach demonstrates improved performance as compared to previous non-zero-shot approaches. \citet{upadhyayjoint} argues that the success of zero-shot cross-lingual approaches \cite{crossling-wikification,sil} might be largely originating from a better estimation of mention-entity prior probabilities. Their approach extends \cite{sil} with global context information and incorporation of typing information into context and entity representations (the system learns to predict typing during the training). The authors report a significant drop in performance for zero-shot cross-lingual EL without mention-entity priors, while showing state-of-the-art results with priors. They also show that training on a resource-rich language might be very beneficial for low-resource settings.

The aforementioned techniques of cross-lingual entity linking heavily rely on pre-trained multilingual embeddings for entity ranking. While being effective in settings with at least prior probabilities available, the performance in realistic zero-shot scenarios drops drastically. Along with the recent success of the zero-shot multilingual transfer of large pre-trained language models, this is a motivation to utilize powerful multilingual self-supervised models. \citet{botha2020entity} use the zeros-shot monolingual architecture of \citet{wufacebook,zero-shot} and mBERT \cite{pires-etal-2019-multilingual} to build a massively multilingual EL model for more than 100 languages. Their system effectively selects proper entities among almost 20 million of candidates using a bi-encoder, hard negative mining, and an additional cross-lingual entity description retrieval task. 
The biggest improvements over the baselines are achieved in the zero-shot and few-shot settings, which demonstrates the benefits of training on a large amount of multilingual data.

\subsection{Methods that do not Fit the General Architecture}

There are a few works that propose methods not fitting the general architecture presented in Figures \ref{fig:edpipeline} and \ref{fig:ranking}. \citet{deeptype} rely on the intermediate supplementary task of entity typing instead of directly performing entity disambiguation. They learn a type system in a KG and train an intermediate type classifier of mentions that significantly refines the number of candidates for the final linking model. \citet{onoe2020fine} leverage distant supervision from Wikipedia pages and the Wikipedia category system to train a fine-grained entity typing model. At test time, they use the soft type predictions and the information about candidate types derived from Wikipedia to perform the final disambiguation. The authors claim that such an approach helps to improve the domain independence of their EL system. \citet{kar} consider a classification approach, where each entity is considered as a separate class or a task. They show that the straightforward classification is difficult due to exceeding memory requirements. Therefore, they experiment with multitask learning, where parameter learning is decomposed into solving groups of tasks. \citet{globerson} do not have any encoder components; instead, they rely on contextual and pairwise feature-based scores. They have an attention mechanism for global ED with a non-linear optimization as described in Section \ref{subsubsec:localglobal}.

\subsection{Summary} \label{subsec:summary}

\begin{table*}[htp]
\caption{\textbf{Features of neural EL models.} Neural entity linking models compared according to their architectural features. The description of columns is presented in the \protect \hyperlink{tablecolumns}{\textcolor{blue}{beginning} of Section  \protect \ref{subsec:summary}}.
The footnotes in the table are enumerated in the \protect \hyperlink{tablefootnotes}{\textcolor{blue} {end}} of Section \ref{subsec:summary}.}

\scalebox{0.65}{\small \begin{tabular}[t]{l|c|c|c|c|c|c|c|c}

\bf Model& \bf \multirowcell{Encoder Type} & \bf Global & \bf \multirowcell{MD+\\ED} &  \multirowcell{\bf NIL\\ \bf Pred.}  & \bf \multirowcell{Ent. Encoder \\Source based on}  & \bf \multirowcell{Candidate\\Generation} & \bf \multirowcell{Learning Type\\for Disam.} &  \bf \multirowcell{Cross-\\lingual}\\ \toprule
 

\mycite{sun} & \multirowcell{CNN+Tensor net.} &&& & ent. specific info.& \multirowcell{surface match+aliases}&supervised& \\ \hline 

\mycite{francis2016capturing}  & CNN &\ding{56}$^3$ &&\ding{56}&ent. specific info.&surface match+prior&supervised& \\ \hline 

\mycite{fang} & word2vec-based & \ding{56} & & & relational info.& n/a  &supervised\\ \hline 

\mycite{yamada} & word2vec-based & \ding{56} & & &relational info. &aliases&supervised\\ \hline 
   
\mycite{zwicklbauer} & word2vec-based & \ding{56} & & \ding{56} & \multirowcell{unstructured text + \\ ent. specific info.}&\multirowcell{surface match} &unsupervised$^5$ \\ \hline

\mycite{crossling-wikification} & word2vec-based & \ding{56} & & \ding{56} & unstructured text &prior &supervised& \ding{56} \\\hline 

\mycite{nguyen} & \multirowcell{CNN} & \ding{56} & & \ding{56} &ent. specific info. &\multirowcell{surface match+prior}&supervised& \\\hline 


\mycite{globerson}&n/a&\ding{56}&&&n/a&prior+aliases&supervised& \\ \hline

\mycite{cao} & word2vec-based & \ding{56} & & & relational info. &aliases&\multirowcell{supervised or\\unsupervised}  &\\ \hline 

\mycite{eshel2017named} & \multirowcell{GRU+Atten.} & & & &unstructured text$^1$&\multirowcell{aliases or surface match}& supervised &  \\ \hline 

\mycite{hofmann} & Atten. & \ding{56} & & & unstructured text & \multirowcell{prior+aliases}&supervised\\ \hline 
    
\mycite{moreno} & word2vec-based & \ding{56}$^3$ & & \ding{56} & unstructured text &\multirowcell{surface match+aliases}&supervised \\ \hline 
   
\mycite{gupta} & LSTM & \ding{56}$^3$ & & & ent. specific info. &prior& supervised$^4$& \\ \hline

\mycite{nie2018mention} & LSTM+CNN & \ding{56} &  && ent. specific info. & surface match+prior & supervised &  \\ \hline

\mycite{sorokin2018mixing} & CNN & \ding{56} & \ding{56} & & relational info. &surface match& supervised&  \\ \hline 

\mycite{shahbazi2018joint} & Atten. & \ding{56} & & & unstructured text &prior+aliases& supervised&  \\ \hline 
 
\mycite{le2018} & Atten. & \ding{56} & & & unstructured text &prior+aliases&supervised\\ \hline 
   
\mycite{griffis} & word2vec-based & &&& unstructured text &aliases&unsupervised&\\ \hline 

\mycite{elden} & n/a & \ding{56}&& & relational info. &aliases&supervised &\\ \hline 

\mycite{end2end} & LSTM & \ding{56} & \ding{56} & & unstructured text & \multirowcell{prior+aliases}&supervised \\ \hline 
   
   
\mycite{sil} & \multirowcell{LSTM+Tensor net.} &&&\ding{56} & ent. specific info. & \multirowcell{prior or\\prior+aliases} & \multirowcell{zero-shot}&\ding{56} \\ \hline 

\mycite{upadhyayjoint} & CNN &\ding{56}$^3$&& & ent. specific info.&prior& \multirowcell{zero-shot} & \ding{56}  \\ \hline 

\mycite{cao2018} & Atten. & \ding{56} & & & relational info. & prior+aliases& supervised \\ \hline 

\mycite{deeptype} & n/a & \ding{56}&&& n/a &\multirowcell{prior+type classifier} & supervised& \ding{56} \\ \hline 

\mycite{mueller} & \multirowcell{GRU+Atten.+CNN}&&&&unstructured text$^1$&surface match&supervised& \\ \hline

\mycite{shahbazi2019entity} & ELMo & & & & unstructured text& \multirowcell{prior+aliases\\or aliases}& supervised&  \\ \hline 
  
\mycite{zero-shot} & BERT & &  & &ent. specific info. &BM25& zero-shot & \\ \hline 
   
\mycite{gillick} & FFNN & & & & ent. specific info. &\multirowcell{nearest neighbors}& supervised$^4$  & \\ \hline 
   
\mycite{knowbert}$^2$ & BERT & \ding{56}$^3$ &\ding{56}&\ding{56}& unstructured text &\multirowcell{prior+aliases}&supervised&\\ \hline 
   
\mycite{titov1} & LSTM & &&&ent. specific info. &surface match&\multirowcell{weakly-\\supervised}&\\ \hline

\mycite{titov2} & Atten. & \ding{56}& & & unstructured text &\multirowcell{prior+aliases}&\multirowcell{weakly-\\supervised}&\\ \hline 
   
\mycite{reinforcement} & LSTM & \ding{56}&&&\multirowcell{unstructured text +\\ent. specific info.}&aliases&supervised& \\ \hline 

\mycite{martins} & LSTM & &\ding{56}&\ding{56}&unstructured text&aliases&supervised  \\ \hline 

\mycite{yang2019} & \multirowcell{Atten. or CNN} & \ding{56} & & &\multirowcell{unstructured text or\\ ent. specific. info.}&{prior+aliases}& supervised&  \\ \hline 

\mycite{ijcai2019-740} & CNN & \ding{56} &&& ent. specific info. & prior+aliases & supervised \\ \hline

\mycite{zhou-etal-2019-towards} & n/a & \ding{56} & & & unstructured text & \multirowcell{prior+char.-\\level model} & zero-shot & \ding{56}  \\ \hline

\mycite{broscheit2020investigating} & BERT & &\ding{56}&\ding{56}&n/a&n/a&supervised& \\ \hline 

\mycite{hou-etal-2020-improving} & Atten. & \ding{56} &  &  & \multirowcell{ent. specific info.+\\unstructured text} & prior+aliases & supervised &   \\ \hline

\mycite{onoe2020fine} & \multirowcell{ELMo+Atten.\\+CNN+LSTM}&&&&n/a&\multirowcell{prior or aliases} &supervised$^4$& \\ \hline

\mycite{chen-etal-2020-contextualized} & BERT &  & \ding{56} &  & relational info. & n/a or aliases & supervised & \\ \hline

\mycite{wufacebook} & BERT & & & &ent. specific info.& \multirowcell{nearest neighbors}&\multirowcell{zero-shot} & \\ \hline 


\mycite{debayan} & fastText & & \ding{56} & & relational info. & surface match&supervised&\\ \hline

\mycite{wu-dgcn}&ELMo&\ding{56}&&&\multirowcell{unstructured text+\\relational info.} &\multirowcell{prior+aliases}& supervised \\ \hline

\mycite{fang-www}& BERT & \ding{56} &&&ent. specific info.&\multirowcell{surface match+aliases+\\Google Search}&supervised \\ \hline

\mycite{chen2020improving} & Atten.+BERT & \ding{56} & & & unstructured text & \multirowcell{prior+aliases} & supervised &
\\ \hline

\mycite{botha2020entity} & BERT & &  & & ent. specific info. & nearest neighbors & zero-shot & \ding{56} \\ \hline

\mycite{yao-etal-2020-zero} & BERT & & & & ent. specific info. & BM25 & zero-shot &   \\ \hline

\mycite{li-etal-2020-efficient} & BERT & & \ding{56} &  & ent. specific info. & nearest neighbors & zero-shot &   \\ \hline

\mycite{poerner-etal-2020-e}$^2$ & BERT & \ding{56} & \ding{56} & \ding{56} & relational info. & prior+aliases & supervised &  \\ \hline

\mycite{fu2020design} & M-BERT &  &  & & ent. specific info. & \multirowcell{Google Search \\ Google Maps} & zero-shot & \ding{56} \\ \hline

\mycite{mulang2020evaluating}  & \multirowcell{Atten. or CNN or BERT} & \ding{56} &  &  & relational info. & prior+aliases & supervised &  \\ \hline

\mycite{yamada20} & BERT & \ding{56}&& &unstructured text&\multirowcell{prior+aliases\\or aliases}&supervised& \\  \hline 

\mycite{mrc} & BERT & \ding{56} && \ding{56}&ent. specific info.&\multirowcell{surface match+prior\\ or aliases}&supervised& \\ \hline

\mycite{tang2021bidirectional} & BERT &  &  &  & ent. specific info. & BM25 & zero-shot & \\ \hline

\mycite{de2020autoregressive} & BART & \ding{56} & \ding{56} & & n/a & prior+aliases & supervised & \\  \hline
   


\end{tabular}}  
\label{table:elmodels}
\end{table*}

We summarize design features for neural EL models in Table \ref{table:elmodels} and also links to their publicly available implementations in Table \ref{table:sourcecodes} in Appendix A.
The mention encoders have made a shift to self-attention architectures and started using deep pre-trained models like BERT. The majority of studies still rely on external knowledge for the candidate generation step. There is a surge of models that tackle the domain adaptation problem in a zero-shot fashion. However, the task of zero-shot joint entity mention detection and linking has not been addressed yet. 
It is shown in several works that the cross-encoder architecture is superior compared to models with separate mention and entity encoders. 
The global context is widely used, but there are few recent studies that focus only on local EL.

\hypertarget{tablecolumns}{Each column} in Table \ref{table:elmodels} corresponds to a model feature.
The \textbf{encoder type} column presents the architecture of the mention encoder of the neural entity linking model. It contains the following options:
    \begin{itemize}[itemsep=1mm, parsep=0pt]
        \item n/a -- a model does not have a neural encoder for mentions / contexts. 
        \item CNN -- an encoder based on convolutional layers (usually with pooling).
        \item Tensor net. -- an encoder that uses a tensor network.
        \item Atten. -- means that a context-mention encoder leverages an attention mechanism to highlight the part of the context using an entity candidate.
        \item GRU -- an encoder based on a recurrent neural network and gated recurrent units
        \cite{chung2014empirical}.
        \item LSTM -- an encoder based on a recurrent neural network and long short-term memory cells \cite{hochreiter1997long} (might be also bidirectional).
        \item FFNN -- an encoder based on a simple feedforward neural network.
        \item ELMo -- an encoder based on a pre-trained ELMo model \cite{peters2018deep}.
        \item BERT -- an encoder based on a pre-trained BERT model \cite{bert}.
        \item fastText -- an encoder based on a pre-trained fastText model \cite{bojanowski-etal-2017-enriching}.
        \item word2vec-based -- an encoder that leverages principles of CBOW or skip-gram algorithms \cite{doc2vec,word2vec,Mikolov2013EfficientEO}.
    \end{itemize}

Note that the theoretical complexity of various types of encoders is different. As discussed by \citet{deep-transformer}, complexity per layer of self-attention is $O(n^2 \cdot d)$, as compared to $O(n \cdot d^2)$ for a recurrent layer, and $O(k \cdot n \cdot d^2)$ for a convolutional layer, where $n$ is the length of an input sequence, $d$ is the dimensionality, and $k$ is the kernel size of convolutions. At the same time, the self-attention allows for a better parallelization than the recurrent networks as the number of sequentially executed operations for self-attention requires a constant number of sequentially executed operations of $O(1)$, while a recurrent layer requires $O(n)$ sequential operations. Overall, estimation of the computational complexity of training and inference of various neural networks is certainly beyond the scope of the goal of this survey. The interested reader may refer to~\cite{deep-transformer} and specialized literature on this topic, e.g.~\cite{orponen1994computational,vsima2003general,livni2014computational}.

The \textbf{global} column shows whether a system uses a global solution (see Section \ref{subsubsec:localglobal}). The \textbf{MD+ED} column refers to joint entity mention detection and disambiguation models, where detection and disambiguation of entities are performed collectively (Section \ref{subsubsec:jointedandel}). The \textbf{NIL prediction} column points out models that also label unlinkable mentions. 
The \textbf{entity embedding} column presents which resource is used to train entity representations based on the categorization in Section \ref{subsec:entityrep}, where
        \begin{itemize}[itemsep=1mm, parsep=0pt]
            \item n/a -- a model does not have a neural encoder for entities.
            \item unstructured text -- entity representations are constructed from unstructured text using approaches based on co-occurrence statistics developed originally for word embeddings like word2vec \cite{word2vec}.
            \item relational info. -- a model uses relations between entities in KGs.
            \item ent. specific info. -- an entity encoder uses other types of information, like entity descriptions, types, or categories. 
        \end{itemize}

In the \textbf{candidate generation} column, the candidate generation methods are specified (Section \ref{subsec:candgen}). It contains the following options:
    \begin{itemize}[itemsep=1mm, parsep=0pt]
         \item n/a -- the solution that does not have an explicit candidate generation step (e.g. the method presented by \citet{broscheit2020investigating}).
         \item surface match -- surface form matching heuristics.
         \item aliases -- a supplementary aliases for entities in a KG.
         \item prior -- filtering candidates with pre-calculated mention-entity prior probabilities or frequency counts.
         \item type classifier -- \citet{deeptype} filter candidates using a classifier for an automatically learned type system.
         \item BM25 -- a variant of TF-IDF to measure similarity between a mention and a candidate entity based on description pages.
         \item nearest neighbors -- the similarity between mention and entity representations is calculated, and entities that are nearest neighbors of mentions are retrieved as candidates. \citet{wufacebook} train a supplementary model for this purpose.
         \item Google search -- leveraging Google Search Engine to retrieve entity candidates.
         \item char.-level model -- a neural character-level string matching model.
    \end{itemize}

The \textbf{learning type for disambiguation} column shows whether a model is \textit{`supervised', `unsupervised', `weakly-supervised', or `zero-shot'}. The \textbf{cross-lingual} column refers to models that provide cross-lingual EL solutions (Section \ref{subsubsec:cross-lingual}). 

\hypertarget{tablefootnotes}{In addition}, the following superscript notations are used to denote specific features of methods shown as a note in the Table \ref{table:elmodels}: 

\begin{enumerate}[itemsep=1mm, parsep=0pt]
\item These works use only entity description pages, however, they are labeled as the first category (unstructured text) since their training method is based on principals from word2vec.
\item The authors provide EL as a subsystem of language modeling. 
\item These solutions do not rely on global coherence but are marked as ``global'' because they use document-wide context or multiple mentions at once for resolving entity ambiguity. 
 \item These studies are domain-independent as discussed in Section \ref{subsubsec:domainindependent}.
 \item \citet{zwicklbauer} may not be accepted as purely unsupervised since they have some threshold parameters in the disambiguation algorithm tuned on a labeled set.
 
\end{enumerate}


\begin{table*}[!ht]

\centering

\footnotesize

\caption{\textbf{Evaluation datasets.} Descriptive statistics of the evaluation datasets used in this survey to compare the EL models. The values for MSNBC, AQUAINT, and ACE2004 datasets are based on the update by \protect \citet{guo18}. The statistics for AIDA-B, MSNBC, AQUAINT, ACE2004, CWEB, and WW is reported according to \protect \cite{hofmann} ($\#$ of mentions takes into account only non-NIL entity references). The TAC KBP dataset statistics is reported according to \protect \cite{tac2010,wufacebook,ellis2015overview,crosslingoverview} ($\#$ of mentions takes into account also NIL entity references).}

\begin{threeparttable}

\begin{tabular}{l|c|c|c} \bf Corpus & \bf Text Genre & \bf $\#$ of Documents & \bf $\#$ of Mentions \\ \toprule    
AIDA-B \cite{hoffart} & News &231 & 4,485\\ \hline
MSNBC \cite{cucerzan2007}& News &20 &656 \\ \hline
AQUAINT \cite{milne}& News &50&727  \\\hline
ACE2004 \cite{ratinov}& News &36&257\\\hline

CWEB \cite{guo18,clueweb} & Web \& Wikipedia & 320 & 11,154\\\hline
WW \cite{guo18} & Web \& Wikipedia & 320 & 6,821\\\hline
TAC KBP 2010 \cite{tac2010} & News \& Web & 2,231 & 2,250 \\ \hline 
\multirowcell{TAC  KBP  2015 Chinese \cite{crosslingoverview}} & \multirowcell{News \& Forums} & 166 & 11,066 \\ \hline
\multirowcell{TAC  KBP  2015 Spanish \cite{crosslingoverview}} & \multirowcell{News \& Forums} & 167 & 5,822 \\ \hline
\end{tabular}


\end{threeparttable}
\label{table:datasetsinfo}


\end{table*}

\begin{table*}[htp]
\footnotesize
\centering
\caption{\textbf{Entity disambiguation evaluation.} Micro F1/Accuracy scores of neural entity disambiguation as compared to some classic models on common evaluation datasets.}
\scalebox{0.87}{\begin{tabular}{l|c|c|c|c|c|c|c|c|c}  \bf Model & \bf{AIDA-B}&

\bf KBP'10 & \bf MSNBC &  \bf AQUAINT &  \bf ACE-2004  &  \bf  CWEB  &  \bf  WW &   \bf  KBP'15 (es) & \bf  KBP'15 (zh)\\ \toprule 
 & Accuracy  & Accuracy & Micro F1 & Micro F1 & Micro F1 & Micro F1 & Micro F1 & Accuracy & Accuracy\\ \toprule

\multicolumn{10}{c}{\bf{Non-Neural Baseline Models}}\\ \toprule

DBpedia Spotlight (\citeyear{mendes}) \cite{mendes}&0.561 &-& 0.421 & 0.518 &0.539& - & - & - & - \\ \hline
AIDA (\citeyear{hoffart}) \cite{hoffart}&0.770&-&0.746&0.571&0.798& - & - & - & -  \\ \hline
\mycite{ratinov}&-&-&0.750&0.830&0.820&0.562&0.672&-&-\\ \hline
WAT (\citeyear{wat}) \cite{wat}&0.805&-&0.788&0.754&0.796& - & - & - & - \\ \hline
Babelfy (\citeyear{naviglielwsd}) \cite{naviglielwsd}&0.758&-&0.762&0.704&0.619& - & - & - & -  \\ \hline
\mycite{lazic} & 0.864 & - & - & - & - & - & - & - & -\\ \hline
\mycite{chisholm} & 0.887 & - & - & - & - & - & - & - & -\\ \hline
PBOH (\citeyear{pboh}) \cite{pboh} & 0.804 & - & 0.861 & 0.841 & 0.832 & - & - & - & -\\\hline

\mycite{guo18} & 0.890 & - & 0.920 & 0.870 & 0.880 & 0.770 & 0.845 & - & - \\ 

\toprule
\multicolumn{10}{c}{\bf{Neural Models}}\\ \toprule
\mycite{sun} & -  & 0.839 & - & - & - & - & - & - & -\\\hline

\mycite{francis2016capturing} & 0.855 & - & - & - & - & - & - & - & -\\\hline

\mycite{fang} & - & 0.889 & 0.755 & 0.852 &  0.808 & - & - & - & -\\\hline
\mycite{yamada} & 0.931 & 0.855 & - & - & - & - & - & - & - \\ \hline
\mycite{zwicklbauer} & 0.784  & - & 0.911 & 0.842 & 0.907 & - & - & - & -\\\hline

\mycite{crossling-wikification} & - & - & - & - & - & - & - & 0.824 & 0.851\\\hline

\mycite{nguyen}&0.872&-&-&-&-&-&-&-&- \\ \hline

\mycite{globerson} &0.927&0.872& - & - & - & - & - & - & - \\ \hline

\mycite{cao} & 0.851 & - & - & - & - & - & - & - & -\\ \hline 

\mycite{eshel2017named} & 0.873  & - & - & - & - & - & - & - & -\\\hline

\mycite{hofmann} & 0.922 & - & 0.937 & 0.885 & 0.885 & 0.779 & 0.775 & - & - \\ \hline 


\mycite{gupta} & 0.829 & - & - & - & 0.907 & - & - & - & -\\\hline

\mycite{nie2018mention} & 0.898 & 0.891 & - & - & - & - & - & - & - \\ \hline


\mycite{shahbazi2018joint} & 0.944 & 0.879 & - & - & - & - & - & - & - \\ \hline 

\mycite{le2018} & 0.931 & - & 0.939 & 0.884 & 0.900 & 0.775 & 0.780 & - & - \\ \hline


\mycite{elden} & 0.930 & 0.896 & - & - & - & - & - & - & -\\\hline

\mycite{end2end}  & 0.831 & - & 0.864 & 0.832 & 0.855 & - & - & - & -\\ \hline

\mycite{sil} & 0.940 & 0.874 & - & - & - & - & - & 0.823 & 0.844\\ \hline

\mycite{upadhyayjoint}  & - & - & - & - & - & - & - & \bf 0.844 & \bf 0.860\\ \hline

\mycite{cao2018} & 0.800 & 0.910 & - & 0.870 & 0.880 & - & 0.860 & - & -    \\\hline

\mycite{deeptype} & 0.949 & 0.909 & - & - & - & - & - & - & -    \\\hline


\mycite{shahbazi2019entity} & 0.962 & 0.883 & 0.923 & 0.901 & 0.887 & 0.784 & 0.798 & - & -\\\hline


\mycite{gillick} & - & 0.870 & - & - & - & - & - & - & -\\\hline
\mycite{titov1} & 0.815 & - & - & - & - & - & - & - & -\\ \hline
\mycite{titov2}  & 0.897 & - & 0.922 & 0.907 & 0.881 & 0.782 & 0.817 & - & -\\ \hline
\mycite{reinforcement}  & 0.943 & - & 0.928 & 0.875 & 0.912 & 0.785 & 0.828 & - & -\\\hline

\mycite{yang2019} & 0.946 & - & 0.946 & 0.885 & 0.901 & 0.756 & 0.788 & - & -\\\hline 

\mycite{ijcai2019-740} & 0.924 &&0.944& 0.919& 0.911&0.801&0.855&-&-\\\hline 


\mycite{zhou-etal-2019-towards} & - & - & - & - & - & - & - & 0.829 & 0.855 \\ \hline

\mycite{hou-etal-2020-improving}& 0.926 & - & 0.943 & 0.912 & 0.907 & 0.785 & 0.819 & - & - \\ \hline

\mycite{onoe2020fine} & 0.859 & - & - & - & - & - & - & - & - \\ \hline
\mycite{wufacebook} & - &  \bf 0.945 & - & - & - & - & - & - & -\\\hline

\mycite{wu-dgcn} & 0.931 & - & 0.927 & 0.894 & 0.906 & \bf 0.814 & 0.792 & - & -\\\hline

\mycite{fang-www} & 0.830 & - & 0.800 & 0.880 & 0.890 & - & - & - & - \\\hline

\mycite{chen2020improving} & 0.937 & - & 0.945 & 0.898 & 0.908 & 0.782 & 0.810 & - & - \\ \hline

\mycite{mulang2020evaluating} & 0.949 & - & - & - & - & - & - & - & - \\\hline 
\mycite{yamada20} & \bf 0.971 & - & \bf 0.963 & \bf 0.935 & \bf 0.919 & 0.789 & \bf 0.892 & - & -\\ \hline
\mycite{de2020autoregressive}& 0.933 & - & 0.943 & 0.909 & 0.911 & 0.773 & 0.879 & - & -\\\hline
\end{tabular}}

\label{table:elresults}
\end{table*}

\section{Evaluation} \label{sec:evaluation}

In this section, we present evaluation results for the entity linking and entity relatedness tasks on the commonly used datasets.

\subsection{Entity Linking}

\subsubsection{Experimental Setup}

The evaluation results are reported based on two different evaluation settings. The first setup is entity disambiguation (ED) where the systems have access to the mention boundaries. The second setup is entity mention detection and disambiguation (MD+ED) where the input for the systems that perform MD and ED jointly is only plain text. We presented their results in separate tables since the scores for the joint models accumulate the errors made during the mention detection phase.

\paragraph{Datasets}

We report the evaluation results of monolingual EL models on the English datasets widely-used in recent research publications: AIDA \cite{hoffart}, TAC KBP 2010 \cite{tac2010}, MSNBC \cite{cucerzan2007}, AQUAINT \cite{milne}, ACE2004 \cite{ratinov}, CWEB \cite{guo18,clueweb}, and WW \cite{guo18}. AIDA is the most popular dataset for benchmarking EL systems.
 For AIDA, we report the results calculated for the test set (AIDA-B).
 

The cross-lingual EL results are reported for the TAC KBP 2015 \cite{crosslingoverview} Spanish (es) and Chinese (zh) datasets. The descriptive statistics of the datasets and their text genres are presented in Table \ref{table:datasetsinfo} according to information reported in \cite{hofmann,wufacebook,tac2010,crosslingoverview,ellis2015overview}.


\begin{figure*}
    \centering
    \includegraphics[width=0.99\textwidth]{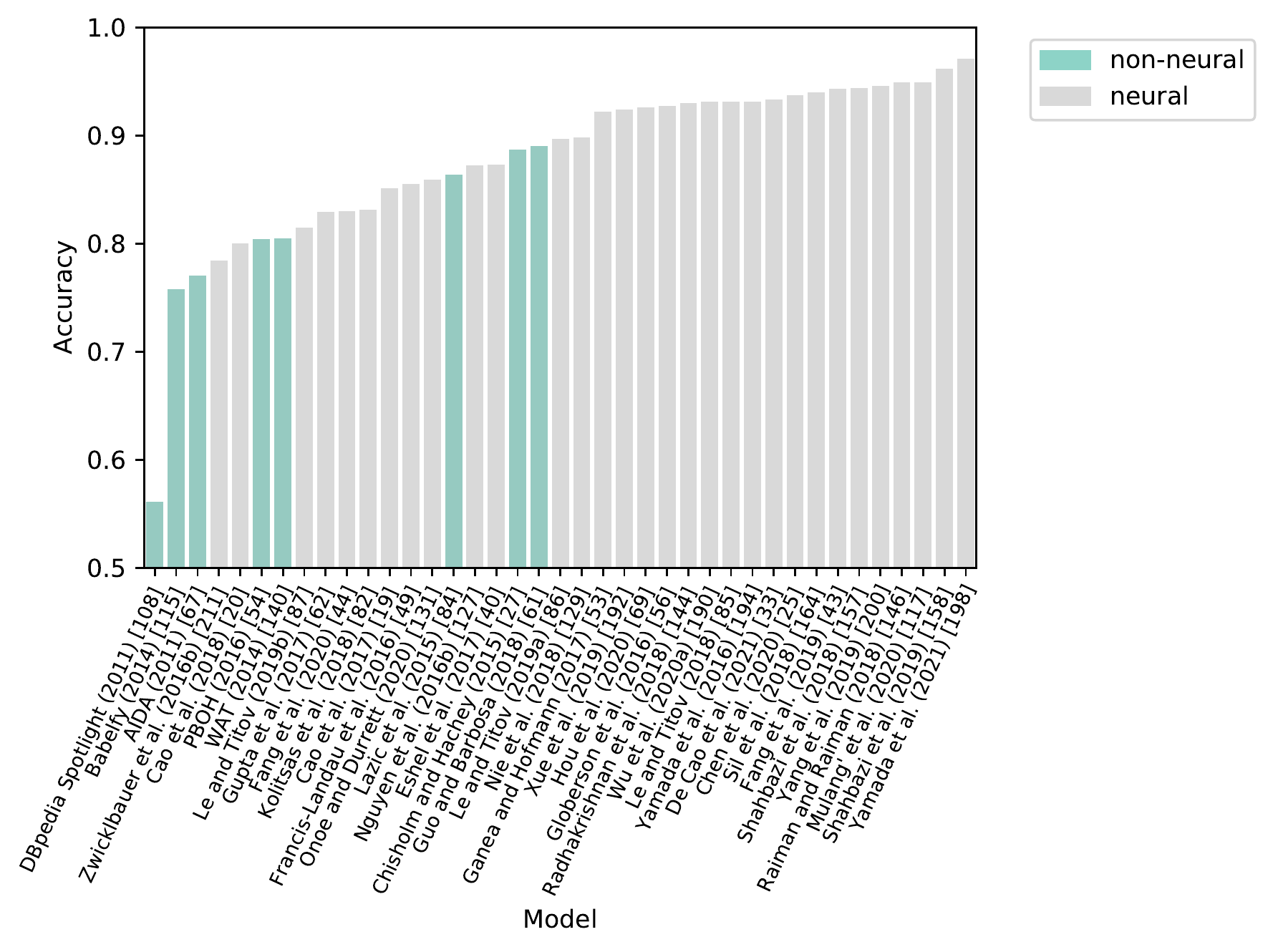}
    \caption{\textbf{Entity disambiguation progress}. Performance of the classic entity linking models (green) with the more recent neural models (gray) on the AIDA test set shows an improvement (around 10 points of accuracy).}
    \label{fig:progress}
\end{figure*}

\begin{figure*}
    \centering
    \includegraphics[width=0.99\textwidth]{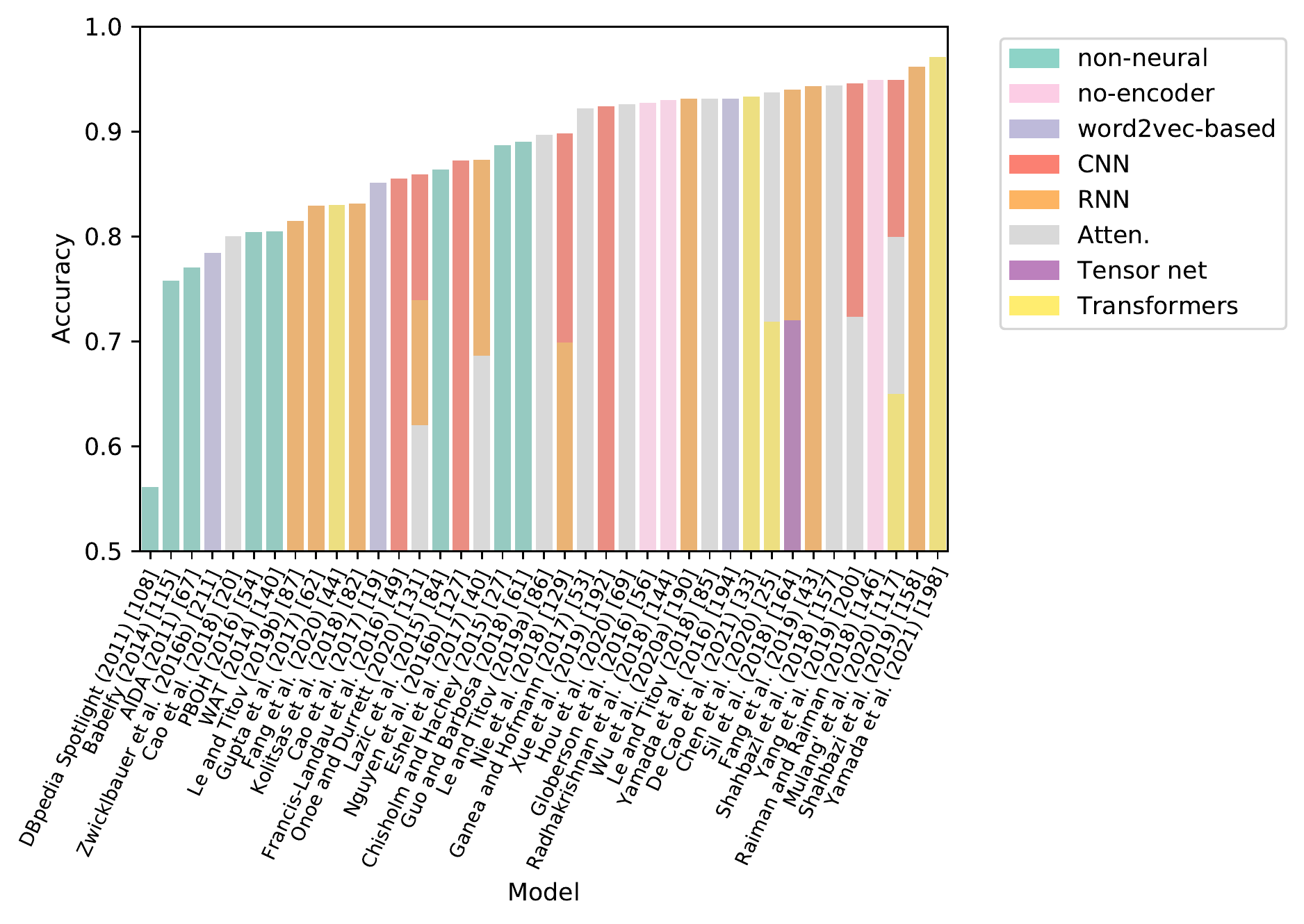}
    \caption{\textbf{Mention/context encoder type for entity disambiguation}. Performance of the entity disambiguation models on the AIDA test set with mention/context encoder displayed with different colors as defined in Table \ref{table:elmodels}. The bars with multiple colors refer to the models that use different types of encoder models; the bars do not reflect any meaning on the percentage. Note: we assigned the ``RNN'' label for the models LSTM, GRU, and ELMo; the ``Transformers'' label for BERT and BART models.}
    \label{fig:mentionencoderprogress}
\end{figure*}


\begin{figure*}
    \centering
    \includegraphics[width=0.99\textwidth]{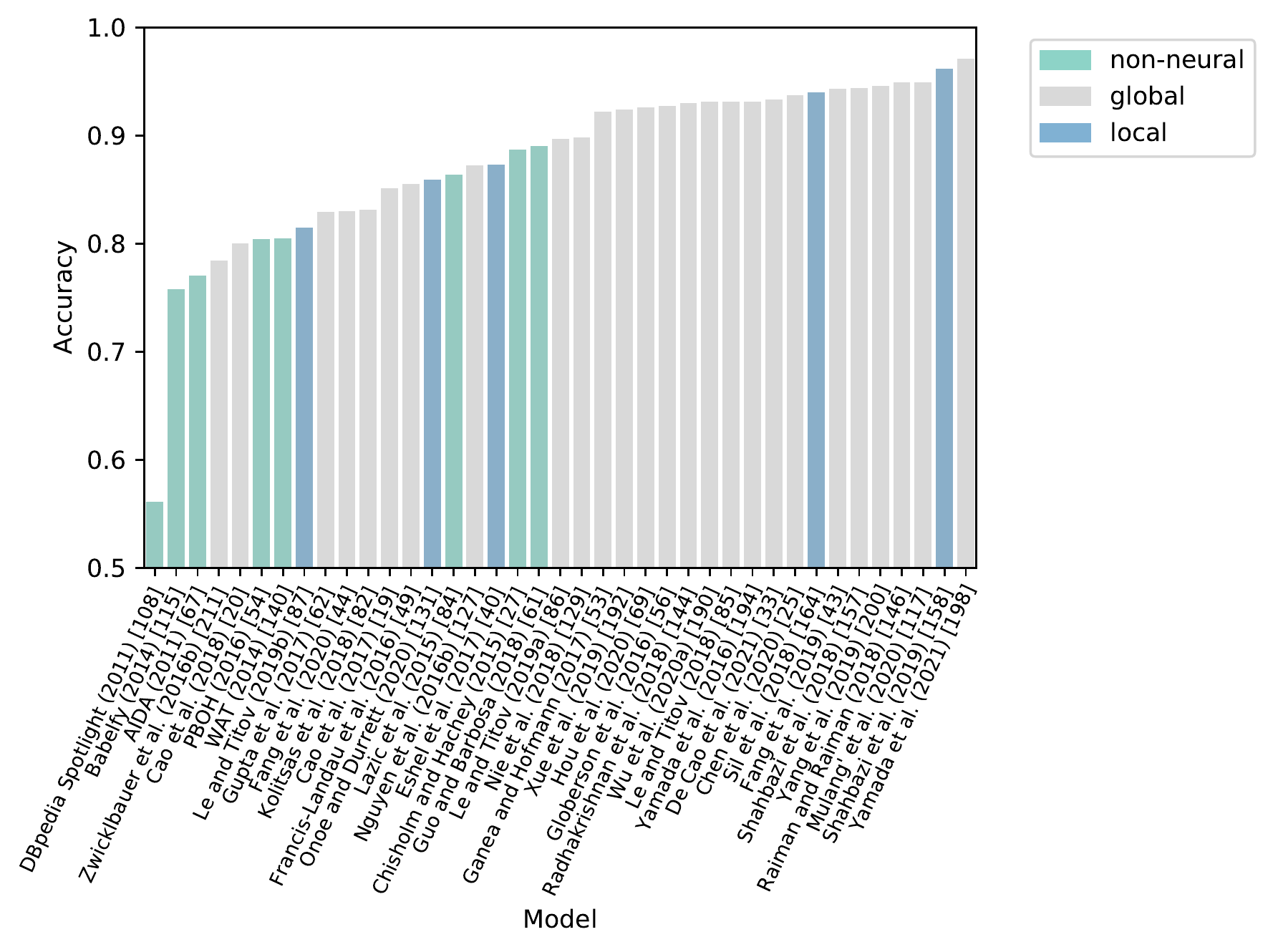}
    \vspace{-0.4cm} 
    \caption{\textbf{Local-Global entity disambiguation}. Performance of the entity disambiguation models on the AIDA test set with local/global models displayed with different colors as defined in Table \ref{table:elmodels}. Note, some models, like \citet{francis2016capturing}, do not rely on global coherence, but they use document-wide context or multiple mentions at once, as explained in Table \ref{table:elmodels}.}
    \label{fig:localglobalprogress}
\end{figure*}

\paragraph{Evaluation Metrics}

For the ED setting, we present micro F1 or accuracy scores reported by model authors. We note that, since mentions are provided as an input, the number of mentions predicted by the model is equal to the number of mentions in the ground truth \cite{surveywei}, so micro F1, precision, recall, and accuracy scores are equal in this setting as explained in \citet{surveywei}:
\begin{equation}
F1 = Acc = \frac{\#\ correctly\ disamb.\ mentions}{\#\ total\ mentions}.
\end{equation}

For the MD+ED setting, where joint models are evaluated, we report micro F1 scores based on strong annotation matching. The formulas to compute F1 scores are shown below, as described in \citet{surveywei} and \citet{pboh}:

\begin{gather}
P = \frac{\#\ correctly\ detected\ and\  disamb.\  mentions}{\#\ predicted\ mentions\ by\ model},  \\
R = \frac{\#\ correctly\ detected\ and\ disamb.\ mentions}{\#\ mentions\ in\ ground\ truth}, \\
F1 = \frac{2 \cdot P  \cdot  R}{P+R}.
\end{gather}
   

We note that results reported in multiple considered papers are usually obtained using GERBIL \cite{gerbil} -- a platform for benchmarking EL models. It implements various experimental setups, including entity disambiguation denoted as D2KB and a combination of mention detection and disambiguation denoted as A2KB. GERBIL encompasses many evaluation datasets in a standartized way along with annotations and provides the computation of evaluation metrics, i.e. micro-\/macro precision, recall, and F-measure.

\paragraph{Baseline Models}

While our goal is to perform a survey of neural EL systems, we also report results of several indicative and prominent classic non-neural systems as baselines to underline the advances yielded by neural models. More specifically, we report results of DBpedia Spotlight (\citeyear{mendes}) \cite{mendes}, AIDA (\citeyear{hoffart}) \cite{hoffart}, \mycite{ratinov}, WAT (\citeyear{wat}) \cite{wat}, Babelfy (\citeyear{naviglielwsd}) \cite{naviglielwsd}, \mycite{lazic}, Chisholm and Hachey (2015) \cite{chisholm}, and PBOH (\citeyear{pboh}) \cite{pboh}. 

For considered neural EL systems, we present the best scores reported by the authors. For the baseline systems, the results are reported according to \citet{end2end}\footnote{Some of the baseline scores are presented in the appendix of \cite{end2end}, which is available at \url{https://arxiv.org/pdf/1808.07699.pdf}} and \citet{hofmann}.


\subsubsection{Discussion of Results}






\paragraph{Entity Disambiguation Results}

We start our discussion of the results from the entity disambiguation (ED)  models, for which mention boundaries are provided. Figure~\ref{fig:progress} shows how the performance of the entity disambiguation models on the most widely-used dataset AIDA improved during the course of the last decade and how the best disambiguation models based on classical machine learning methods (denoted as ``non-neural'') correspond to the recent state-of-the-art models based on deep neural networks (denoted as ``neural''). As one may observe, the models based on deep learning substantially improve the EL performance pushing the state of the art by around 10 percentage points in terms of accuracy.

Table \ref{table:elresults} presents the comparison of the ED models in detail on several datasets presented above. 
The model of \citet{yamada20} yields the best result on AIDA and appears to behave robustly across different datasets, getting top scores or near top scores for most of them. Here, we should also mention that none of the non-neural baselines reach the best results on any dataset. 

Among local models for disambiguation, the best results are reported by  \citet{shahbazi2019entity} and \citet{wufacebook}. It is worth noting that the latter model can be used in the zero-shot setting. \citet{shahbazi2019entity} has the best score on AIDA among other local models outperforming them by a substantial margin. However, this is due to the use of the less-ambiguous resource of \citet{pershina} for candidate generation, while many other works use the YAGO-based resource provided by \citet{hofmann}, which typically yields lower results.

The common trend is that the global models (those trying to disambiguate several entity occurrences at once) outperform the local ones (relying on a single mention and its context). The best considered ED model of \citet{yamada20} is global. 
Its performance improvements over competitors are attributed by the authors to the novel masked entity prediction objective that helps to fine-tune pre-trained BERT for producing contextualized entity embeddings and to the multi-step global disambiguation algorithm.

Finally, as one could see from Table \ref{table:elresults}, the least number of experiments is reported on the non-English datasets (TAC KBP datasets for Chinese and Spanish). Among the four reported results, the approach of \citet{upadhyayjoint} provides the best scores, yet outperforming the other three approaches only by a small margin.

\paragraph{Mention/Context Encoder Type}

Figure \ref{fig:mentionencoderprogress} provides further analysis of the performance of entity disambiguation models presented above. The top performing model by \citet{yamada20} is based on Transformers. It is followed by the model of \citet{shahbazi2019entity}, which relies on RNNs: more specifically, it relies on the ELMo encoder that is based on pre-trained bidirectional LSTM cells. Overall, RNN is a popular choice for the mention-context encoder. However, recently, self-attention-based encoders, and especially the ones based on pre-trained Transformer networks, have gained popularity.

Several approaches, such as \citet{yamada}, rely on simpler encoders based on the word2vec models, yet none of them manage to outperform more complex deep architectures.


\paragraph{Local-global models}

Figure \ref{fig:localglobalprogress} visualizes the usage of the local and global context in various models for entity disambiguation. As one can observe from the plot, the majority of models perform global entity disambiguation, including the top-performing model by \citet{yamada20}. Although \citet{shahbazi2019entity} provide a local model, they also show a good performance.

\begin{table}
\footnotesize
\centering
\caption{\textbf{Evaluation of joint MD-ED models.} Micro F1 scores for joint entity mention detection and entity disambiguation evaluation on AIDA-B and MSNBC datasets.
}
\scalebox{1.09}{\begin{tabular}{l|c|c} \bf Model & \bf{AIDA-B} & \bf{MSNBC}\\ \toprule  
\multicolumn{3}{c}{\bf{Non-Neural Baseline Models}}\\ \toprule
DBpedia Spotlight (\citeyear{mendes}) \cite{mendes} & 0.578 & 0.406 \\ \hline
AIDA (\citeyear{hoffart}) \cite{hoffart} & 0.728&0.651\\ \hline
WAT (\citeyear{wat}) \cite{wat} & 0.730&0.645\\ \hline
Babelfy (\citeyear{naviglielwsd}) \cite{naviglielwsd}&0.485&0.397 \\ \hline\toprule
\multicolumn{3}{c}{\bf{Neural Models}}\\ \toprule
\mycite{end2end} & 0.824 & 0.724\\ \hline
\mycite{martins} & 0.819 & -\\\hline
\mycite{knowbert} & 0.744 & -\\ \hline 
\mycite{broscheit2020investigating} & 0.793 & - \\ \hline
\mycite{chen-etal-2020-contextualized} & \bf 0.877 & - \\ \hline
\mycite{poerner-etal-2020-e} & 0.850 & - \\ \hline
\mycite{de2020autoregressive} &  0.837 & \bf 0.737 \\ \hline
\end{tabular}}

\label{table:elresultsjoint}
\end{table}

\paragraph{Joint Entity Mention Detection and Disambiguation}

Table \ref{table:elresultsjoint} presents results of the joint MD and ED models. Only a fraction of the models presented in Table \ref{table:elmodels} is capable of performing both entity mention detection and disambiguation; thus, the list of results is much shorter. Among the joint MD and ED solutions, the best results on the AIDA dataset are reported by \citet{chen-etal-2020-contextualized}. However, \citet{poerner-etal-2020-e} note that these results might not be directly comparable with others due to a different evaluation protocol. The best comparable results on the AIDA dataset are shown by E-BERT \cite{poerner-etal-2020-e}. On the MSNBC dataset, the top scores are achieved by \citet{de2020autoregressive} with an autoregressive model. The scores of the systems that solve both tasks at once fall behind the disambiguation-only systems since they rely on noisy mention boundaries produced by themselves. In the joint MD and ED setting, the neural models also substantially (up to around 10 points) outperform the classic models. 


\paragraph{On Effect of Hyperparameter Search} 

As explained above, in Tables \ref{table:elresults} and \ref{table:elresultsjoint}, we present the best scores reported by the authors of the models. In principle, each neural model can be further tuned as shown by \citet{reimers-gurevych-2017-reporting}, but also the variance of neural models is rather high in general. Therefore, it may be possible to further optimize meta-parameters of one (possibly simpler) neural model so that it outperforms a more complex (but tuned in a less optimal way) model. One common example of such a case is RoBERTa~\cite{liu2019roberta}, which is basically the original BERT model, which was carefully and robustly optimized. This model outperformed many successors of the BERT model, showing the new state-of-the-art results on various tasks while keeping the original architecture.

\subsection{Entity Relatedness}

The quality of entity representations can be measured by how they capture semantic relatedness between entities \cite{huang,hofmann,yamada,cao,shi}. Moreover, the semantic relatedness is an important feature in global EL \cite{elvaigh,ceccarelli}.
In this section, we present results of entity relatedness evaluation, which is different from evaluation of EL pipelines.

\begin{table*}
\footnotesize
\centering
\caption{\textbf{Entity relatedness evaluation.} Reported results for entity relatedness evaluation on the test set of \protect \citet{ceccarelli} .}
\begin{tabular}{l|c|c|c|c}

 \bf Model &  \bf  nDCG@1 &  \bf  nDCG@5 &  \bf  nDCG@10 &  \bf MAP  \\ \toprule

\mycite{milne} & \textit{0.540} & \textit{0.520} & \textit{0.550} & \textit{0.480} \\ \hline 

\mycite{huang} & \bf 0.810 & 0.730 & 0.740 & \bf 0.680 \\ \hline 

\mycite{yamada} & 0.590 & 0.560 & 0.590 & 0.520 \\ \hline 

\mycite{hofmann} & 0.632 & 0.609 & 0.641 & 0.578\\ \hline

\mycite{cao} & 0.613 & 0.613 & 0.654 & 0.582\\ \hline

\mycite{elvaigh} & \textit{0.690} & \textit{0.640} & \textit{0.580} & -\\ \hline

\mycite{shi} & 0.680 & \bf 0.814 & \bf 0.820 & -\\  \hline
\end{tabular}

\label{table:resultembed}
\end{table*}

\subsubsection{Experimental Setup}


We summarize results from several works obtained on a benchmark of \citet{ceccarelli} for entity relatedness evaluation based on the dataset of \citet{hoffart}.
Given a target entity and a list of candidate entities, the task is to rank candidates semantically related to the target higher than the others \cite{hofmann}. 
For the most of the considered works, the relatedness is measured by the cosine similarity of entity representations. For comparison, we also add results for two other approaches: a well-known Wikipedia hyperlink-based measure devised by \citet{milne} known as WLM and a KG-based measure of \citet{elvaigh}.


The evaluation metrics are normalized discounted cumulative gain (nDCG) \cite{ndcg} and a mean average precision (MAP) \cite{map}. nDCG is a commonly used metric in information retrieval. It discounts the correct answers, depending on their rank in predictions \citet{map}:
%
\begin{equation}
    nDCG(Q, k) = \frac{1}{|Q|}\sum_{j=1}^{|Q|} Z_{kj} \sum_{m=1}^{k} \frac{2^{R(j,m)}-1}{{\log_{2}(1+m)}},
\label{eq:nDCG}
\end{equation}

where $Q$ is the set of target entities (queries); $Z_{kj}$ is a normalization factor, which corresponds to ideal ranking; $k$ is a number of candidates for each query; $R(j,m) \in \{0,1\}$ is the gold-standard annotation of relatedness between the target entity $j$ and a candidate $m$. 



MAP is another common metric in information retrieval \cite{map}:
\begin{equation}
    MAP(Q) = \frac{1}{|Q|}\sum_{j=1}^{|Q|} \frac{1}{m_j}\sum_{k=1}^{m_j} Precision@r_{jk},
\label{eq:MAP}
\end{equation}

where $Q$ is a set of target entities (queries); $m_j$ is the number of related candidate entities for the target $j$, and $Precision@r_{jk}$ is a precision at rank $r_{jk}$, where $r_{jk}$ is a rank of each related candidate in the prediction $k=1..m_j$ \cite{map}.  

\subsubsection{Discussion of Results}

Table \ref{table:resultembed} summarizes the evaluation results in the entity relatedness task reported by the authors of the models. The scores of \citet{milne} are taken from \citet{huang}.

The highest scores of nDCG@1 and MAP are reported by \citet{huang}, and the best scores of nDCG@5 and nDCG@10 are reported by \citet{shi}. The high scores of \citet{huang} can be attributed to the usage of different information sources for constructing entity representations, including entity types and entity relations \cite{hofmann}.
\citet{shi} also use various types of data sources for constructing entity representations, including textual and knowledge graph information, like the types provided by a category hierarchy of a knowledge graph. 

Note that cosine similarity based measures perform better in terms of nDCG@10 than the methods based on relations in KG (shown as italic in Table~\ref{table:resultembed}).


\section{Applications of Entity Linking}
\label{sec:applications}

In this section, we first give a brief overview of established applications of the entity linking technology and then discuss recently emerged use-cases specific to neural entity linking based on injection of these models as a part of a larger neural network, e.g. in a neural language model. 


\subsection{Established Applications}

\paragraph{Text Mining}

An EL tool is a typical building block for text mining systems. Extracting and resolving the ambiguity of entity mentions is one of the first steps in a common information extraction pipeline. The ambiguity problem is especially crucial for such domains as biomedical and clinical text processing due to variability of medical terms, the complexity of medical ontologies such as UMLS \cite{umls}
, and scarcity of annotated resources. There is a long history of development of EL tools for biomedical literature and electronic health record mining applications \cite{aronson2010overview,savova2010mayo,soldaini2016quickumls,bertformedel2020,zhu2020latte,medlinker,medcat,biomedicalel2021,chen2021lightweight}. These tools have been successfully applied for summarization of clinical reports \cite{macavaney2019ontology}, extraction of drug-disease treatment relationships \cite{khare2014labeledin}, mining chemical-induced disease relations~\cite{bansal2020simultaneously}, differential diagnosis \cite{amiri-etal-2021-attentive}, patient screening \cite{eyre2021launching}, and many other tasks. Besides medical text processing, EL is widely used for mining social networks and news \cite{moon-etal-2018-multimodal-named,multimodaleltweets2020}. For example, Twitcident \cite{abel2012twitcident} uses the DBpedia Spotlight \cite{mendes} EL system for mining Twitter messages for small scale incidents. \citet{provatorova2020named} leverage a recently proposed EL toolkit REL \cite{rel} for mining historical newspapers for people, places, and other entities in the CLEF HIPE 2020 evaluation campaign \cite{ehrmann2020overview}. \citet{luo2021newsclippings} automatically construct a large-scale dataset of images and text captions that describe real and out-of-context news. They leverage REL for linking entities in image captions, which helps to automatically measure inconsistency between images and their text captions.

\paragraph{Knowledge graph population}

EL is one of the necessary steps of knowledge graph population algorithms. 
Before populating a KG with new facts extracted from raw texts, we have to determine mentioned concepts in these texts and link them to the corresponding graph nodes. A series of evaluation workshops TAC\footnote{https://tac.nist.gov/2019/index.html} provides a forum for KG population tools (TAC KBP), as well as benchmarks for various subsystems including EL. For example, \citet{ji2011knowledge} and \citet{ellis2015overview} overview various successful systems for knowledge graph population participated in the TAC KBP 2010 and 2015 tasks. \citet{shineplus} propose a knowledge graph population algorithm that not only uses the results of EL, but also helps to improve EL itself. It iteratively populates a KG, while the EL model benefits from added knowledge and continuously learns to disambiguate better.

%

\paragraph{Information retrieval and question-answering}

EL is also widely used in information retrieval and question-answering systems. EL helps to complement search results with additional semantic information, to resolve query ambiguity, and to restrict the search space. For example, \citet{biosearchengine} use EL to complement the results of a biomedical literature search engine with found entities: genes, diseases, drugs, etc. COVIDASK \cite{covid19}, a real-time question answering system that helps researchers to retrieve information related to coronavirus, uses the BioSyn model \cite{biomedical} for processing COVID-19 articles and linking mentions of drugs, symptoms, diseases to concepts in biomedical ontologies. Links to entity descriptions help users to navigate the search results, which enhances the usability of the system.
\citet{yih2015semantic} apply EL for pruning the search space of a question answering system. For the query: ``Who first voiced Meg on Family Guy?'', after linking ``Meg'' and ``Family Guy'' to entities in a KG, the task becomes to resolve the predicates to the ``Family Guy (the TV show)'' entry rather than all entries in the KG. \citet{shnayderman2019fast} develop a fast EL algorithm for pre-processing large corpora for their autonomous debating system \cite{slonim2021autonomous} with the goal to conduct an argumentative dialog with an opponent on some topic and to prove a predefined point of view. The system uses the results of entity linking for corpus-based argument retrieval. 

\subsection{Novel Applications: Neural Entity Linking for Training Better Neural Language Models}
\label{sec:novel}


Neural EL models have unlocked the new category of applications that have not been available for classical machine learning methods. Namely, neural models allow the integration of an entire entity linking system inside a larger neural network such as BERT. As they are both neural networks, such kind of integration becomes possible. After integrating an entity linker into another model's architecture, we can also expand the training objective with an additional EL-related task and train parameters of all neural components jointly:
\begin{equation}
\mathcal{L}_{\text {JOINT }}=\mathcal{L}_{\text {BERT }}+\mathcal{L}_{\text {EL-related }}.
\end{equation}





Neural entity linkers can be integrated in any other networks. The main novel trend is the use of EL information for representation learning. Several studies have shown that contextual word representations could benefit from information stored in KGs by incorporating EL into deep language models (LMs) for transfer learning.

KnowBERT \cite{knowbert}  injects one or several entity linkers between top layers of the BERT architecture and optimizes the whole network for multiple tasks: the masked language model (MLM) task and next sentence prediction (NSP) from the original BERT model, as well as EL:
%
%
\begin{equation}
\mathcal{L}_{\mathrm{BERT}}=\mathcal{L}_{\mathrm{NSP}}+\mathcal{L}_{\mathrm{MLM}}.
\end{equation}
\vspace{-0.7cm}
\begin{equation}
\mathcal{L}_{\text {KnowBert }}=\mathcal{L}_{\mathrm{NSP}}+\mathcal{L}_{\mathrm{MLM}}+\mathcal{L}_{\text {EL }}.
\end{equation}

The authors adopt the general end-to-end EL architecture of \cite{end2end} but use only the local context for disambiguation and an encoder based on self-attention over the representations generated by underlying BERT layers. If the EL subsystem detects an entity mention in a given sentence, corresponding pre-built entity representations of candidates are utilized for calculating the updated contextual word representations generated on the current BERT layer. These representations are used as input in a subsequent layer and can also be modified by a subsequent EL subsystem. Experiments with two EL subsystems based on Wikidata and WordNet show that presented modifications in KnowBERT help it to slightly surpass other deep pre-trained language models in tasks of relationship extraction, WSD, and entity typing.

ERNIE \cite{ernie} expands the BERT \cite{bert} architecture with a knowledgeable encoder (K-Encoder), which fuses contextualized word representations obtained from the underlying self-attention network with entity representations from a pre-trained TransE model \cite{transe}. EL in this study is performed by an external tool TAGME \cite{ferragina2010tagme}. For model pre-training, in addition to the MLM task, the authors introduce the task of restoring randomly masked entities in a given sequence keeping the rest of the entities and tokens. They refer to this procedure as a denoising entity auto-encoder (dEA):
\begin{equation}
\mathcal{L}_{\text {ERNIE }}=\mathcal{L}_{\mathrm{NSP}}+\mathcal{L}_{\mathrm{MLM}}+\mathcal{L}_{\text {dEA }}.
\end{equation}

Using English Wikipedia and Wikidata as training data, the authors show that introduced modifications provide performance gains in entity typing, relation classification, and several GLUE tasks \cite{wang2018glue}.

\citet{wang2019kepler} train a disambiguation network named KEPLER using the composition of two losses: regular MLM and a Knowledge Embedding (KE) loss based on the TransE \cite{transe} objective for encoding graph structures: 
\begin{equation}
\mathcal{L_{\mathrm{KEPLER}}}=\mathcal{L}_{\mathrm{MLM}}+\mathcal{L}_{\mathrm{KE}}.
\end{equation}

In the KE loss, representations of entities are obtained from their textual descriptions encoded with a self-attention network \cite{liu2019roberta}, and representations of relations are trainable vectors. 
The network is trained on a dataset of entity-relation-entity triplets with descriptions gathered from Wikipedia and Wikidata. 
Although the system exhibits a significant drop in performance on general NLP benchmarks such as GLUE \cite{wang2018glue}, it shows increased performance on a wide range of KB-related tasks such as TACRED \cite{zhang2017tacred}, FewRel \cite{han2018fewrel}, and OpenEntity \cite{openentity}.

\citet{yamada-etal-2020-luke} propose a deep pre-trained model called ``Language Understanding with Knowledge-based Embeddings'' (LUKE). They modify RoBERTa \cite{liu2019roberta} by introducing an additional pre-training objective and an entity-aware self-attention mechanism. The objective is a simple adoption of the MLM task to entities $\mathcal{L}_{MLMe}$, instead of tokens, the authors suggest restoring randomly masked entities in an entity-annotated corpus.  
\begin{equation}
\mathcal{L_{\mathrm{LUKE}}}=\mathcal{L}_{\mathrm{MLM}}+\mathcal{L}_{\mathrm{MLMe}}.
\end{equation}

Although the corpus used in this work is constructed from Wikipedia by considering hyperlinks to other Wikipedia pages as mentions of entities in a KG, alternatively, it can be generated using an external entity linker.

The entity-aware attention mechanism helps LUKE differentiate between words and entities via introducing four different query matrices for matching words and entities: one for each pair of input types (entity-entity, entity-word, word-entity, and the standard word-word). The proposed modifications give LUKE exceptional performance improvements over previous models in five tasks: Open Entity (entity typing) \cite{openentity}, TACRED (relation classification) \cite{zhang2017tacred}, CoNLL-2003 (named entity recognition) \cite{tjong-kim-sang-de-meulder-2003-introduction}, ReCoRD (cloze-style question answering) \cite{record}, and SQuAD~1.1 (reading comprehension) \cite{rajpurkar-etal-2016-squad}.

\citet{fevry-etal-2020-entities} propose a method for training a language model and entity representations jointly, which they call Entities as Experts (EaE). The model is based on the Transformer architecture and is similar to KnowBERT \cite{knowbert}. However, in addition to the trainable word embedding matrix, EaE features a separate trainable matrix for entity embeddings referred to as ``memory''. 
The standard Transformer is also extended with an ``entity memory'' layer, which takes the output from the preceding Transformer layer and populates it with entity embeddings of mentions in the text. The retrieved entity embeddings are integrated into token representations by summation before layer normalization. To avoid dependence
at inference on an external mention detector, the model applies a classifier to the output of Transformer blocks as in a sequence labeling model.

Analogously to \cite{yamada-etal-2020-luke}, the EaE is trained on a corpus annotated with mentions and entity links. The final loss function sums up of three components: the standard MLM objective, mention boundary detection loss as in a sequence labeling model $\mathcal{L}_{NER}$, and an entity linking objective that facilitates entity representations generated in the model to be close to entity embedding of an annotated entity.
\begin{equation}
\mathcal{L_{\mathrm{EaE}}}=\mathcal{L}_{\mathrm{MLM}}+\mathcal{L}_{\mathrm{NER}}+\mathcal{L}_{\mathrm{EL}}.
\end{equation}

This approach to integrating knowledge about entities into LMs provides a significant performance boost in open domain question answering. EaE, having only 367 million of parameters, outperforms the 11 billion parameter version of T5 \cite{2020t5} on the TriviaQA task \cite{joshi-etal-2017-triviaqa}. The authors also show that EAE contains more factual knowledge than a comparably-sized BERT model.

\citet{poerner-etal-2020-e} present an E-BERT language model that also takes advantage of entity representations. This model is close to \cite{ernie} as it also injects entities directly into the text and mixes entity representations with word embeddings in a similar way. However, instead of updating the weights of the whole pre-trained language model, they train only a linear transformation for aligning pre-trained entity representations with representations of word piece tokens of BERT. Such a small modification helps this model to outperform baselines on unsupervised question answering, supervised relation classification, and end-to-end entity linking.

The considered works demonstrate that the integration of structured KGs and LMs usually helps to solve knowledge-oriented tasks: question answering (including open-domain QA), entity typing, relation extraction, and others. A high-precision supervision signal from KGs either leads to notable performance improvements or allows to reduce the number of trainable parameters of an LM while keeping a similar performance. Entity linking acts as a bridge between highly structured knowledge graphs and more flexible language models. We expect this approach to be crucial for the construction of future foundation models.






\section{Conclusion} \label{sec:conclusion}

In this survey, we have analyzed recently proposed neural entity linking models, which generally solve the task with higher accuracy than classical methods. We provide a generic neural entity linking architecture, which is applicable for most of the neural EL systems, including the description of its components, e.g. candidate generation, entity ranking, mention and entity encoding. Various modifications of the general architecture are grouped into four common directions: (1) joint entity mention detection and linking models, (2) global entity linking models, (3) domain-independent approaches, including zero-shot and distant supervision methods, and (4) cross-lingual techniques. Taxonomy figures and feature tables are provided to explain the categorization and to show which prominent features are used in each method. 

The majority of studies still rely on external knowledge for the candidate generation step. The mention encoders have made a shift from convolutional and recurrent models to self-attention architectures and start using pre-trained contextual language models like BERT. There is a current surge of methods that tackle the problem of adapting a model trained on one domain to another domain in a zero-shot fashion. These approaches do not need any annotated data in the target domain, but only descriptions of entities from this domain to perform such adaptation. It is shown in several works that the cross-encoder architecture is superior as compared to models with separate mention and entity encoders. The global context is widely used, but there are few recent studies that focus only on local EL. 

Among the solutions that perform mention detection and entity disambiguation jointly, the leadership is owned by the entity-enhanced BERT model (E-BERT) of \citet{poerner-etal-2020-e} and the autoregressive model of \citet{de2020autoregressive} based on BART. 
Among published local models for disambiguation, the best results are reported by \citet{shahbazi2019entity} and \citet{wufacebook}. The former solution leverages entity-aware ELMo (E-ELMo) trained to additionally predict entities along with words as in language-modelling task. The latter solution is based on a BERT bi-/cross-encoder and can be used in the zero-shot setting. 
\citet{yamada20} report results that are consistently better in comparison to all other solutions. Their high scores are attributed to the masked entity prediction mechanism for entity embedding and the usage of the pre-trained model based on BERT with a multi-step global scoring function. 


\section{Future Directions} \label{sec:futuredirections}

We identify five  promising directions of future work in entity linking listed below:  

\begin{enumerate}
\item \textbf{More end-to-end models without an explicit candidate generation step}: 
The candidate generation step relies on pre-constructed external resources or heuristics, as discussed in Section \ref{subsec:candgen}. Both the recall and precision of EL systems depend on their completeness and ambiguity. The necessity of building such resources is also an obvious obstacle for applying models in zero-shot / cross-lingual settings. Several recent works demonstrate that it is possible to achieve high EL performance without external pre-built resources \cite{gillick,wufacebook} or eliminate the candidate generation step  \cite{broscheit2020investigating,botha2020entity}. There is also a line of works devoted to methods that perform mention detection and entity disambiguation jointly \cite{end2end,de2020autoregressive}, which helps to avoid error propagation through multiple independent processing steps in an EL pipeline. We believe that a possible further research direction would be the development of entirely end-to-end trainable EL pipelines similar in spirit to the system of \citet{broscheit2020investigating}.

\item \textbf{Further development of zero-shot approaches to address emerging entities}: We also expect that zero-shot EL will rapidly evolve, engaging other features like global coherence across all entities in a document, NIL prediction, joining MD and ED steps together, or providing completely end-to-end solutions. The latter would be an especially challenging task but also a fascinating research direction. To allow for a proper comparison, more standardized benchmarks and evaluation processes for zero-shot methods are dearly needed.

\item \textbf{More use-cases of EL-enriched language models}: Some studies \cite{knowbert,ernie,wang2019kepler,poerner-etal-2020-e} have shown improvements over contextual language models by including knowledge stored in KGs. They incorporate entity linking into these deep models to use information in KGs. In future work, more use-cases are expected to enhance language models by using entity linking. The enriched representations would be used in downstream tasks, enabling improvements there. 

\item \textbf{Integration of EL loss in more neural models}: It may be interesting to integrate EL loss in other neural models distinct from the language models, but in a similar fashion as the models described in Section~\ref{sec:novel}. Due to the fact that an end-to-end EL model is also just a neural network, such integration with other networks is technically straightforward. Some multi task learning methods have been already proposed, e.g. joint relation extraction and entity linking \cite{bansal2020simultaneously}.
Since entity linking is a key step in information extraction, injecting information about entities contained in an EL model and multitask learning are expected to be useful for solving other related tasks.  

\item \textbf{Multimodal EL}: We witness the rise of a fascinating information extraction research direction that aims to build models capable of processing not only text, but also data from other modalities like images. For example, \citet{moon-etal-2018-multimodal-named} and \citet{multimodaleltweets2020} leverage both text and images in social media posts for entity linking. Without taking into account an additional modality it would be impossible to correctly disambiguate entities in a very noisy and limited textual context. Entity linking methods in the near future potentially could take advantage of multimodal cross-attention and a surge of other techniques recently developed to improve processing multiple types of data in a single architecture \cite{nagrani2021attention,Jaegle2021PerceiverGP}. We consider that vice-versa is also possible: EL could be seamlessly integrated into models for processing data with multiple modalities. EL not only provides disambiguation of mentions in the text but also connects a data instance to a knowledge graph, which opens the possibility of using reasoning elements during the solution of the final task.


\end{enumerate}




\begin{acks}
The work was partially supported by a Deutscher Akademischer Austauschdienst (DAAD) doctoral stipend and the DFG-funded JOIN-T project BI 1544/4. The work of Artem Shelmanov in the current study (preparation of sections related to application of entity linking to neural language models, entity ranking, context-mention encoding, and overall harmonization of the text and results) is supported by the Russian Science Foundation (project 20-11-20166). Finally, this work was partially supported by the joint MTS-Skoltech laboratory.  

\end{acks}

\begin{appendix}
\onecolumn

\section{Public Implementations of Neural Entity Linking Models}

\begin{table*}[!h]
\caption{Publicly available implementations (either provided in the paper or available at  \href{https://paperswithcode.com}{PapersWithCode.com}) of the neural models presented in Table \protect \ref{table:elmodels}.}
\scalebox{0.65}{\begin{tabular}[t]{l|l}
\bf Model  & \bf Link for Source Code \\ \hline  \toprule
 
\mycite{sun} &  - \\ \hline 

\mycite{francis2016capturing}  &\url{https://github.com/matthewfl/nlp-entity-convnet}   \\ \hline 

\mycite{fang} & - \\ \hline 

\mycite{yamada} & \url{https://github.com/wikipedia2vec/wikipedia2vec}  \\ \hline
   
\mycite{zwicklbauer} &\url{https://github.com/quhfus/DoSeR} \\ \hline

\mycite{crossling-wikification} &-  \\\hline 

\mycite{nguyen}  & - \\\hline 

\mycite{globerson} & - \\ \hline

\mycite{cao} & \url{https://github.com/TaoMiner/bridgeGap} \\ \hline 

\mycite{eshel2017named} & \url{https://github.com/yotam-happy/NEDforNoisyText} \\ \hline 

\mycite{hofmann} &\url{https://github.com/dalab/deep-ed} \\ \hline

\mycite{moreno} &-  \\ \hline 

\mycite{gupta} & \url{https://github.com/nitishgupta/neural-el} \\ \hline

\mycite{nie2018mention} & - \\ \hline

\mycite{sorokin2018mixing} & \url{https://github.com/UKPLab/starsem2018-entity-linking}  \\ \hline 

\mycite{shahbazi2018joint} & -  \\ \hline
 
\mycite{le2018} &  \url{https://github.com/lephong/mulrel-nel} \\ \hline 

\mycite{griffis} &  \url{https://github.com/OSU-slatelab/JET} \\ \hline 

\mycite{elden} & \url{https://github.com/priyaradhakrishnan0/ELDEN}  \\ \hline 

\mycite{end2end} & \url{https://github.com/dalab/end2end_neural_el} \\ \hline

\mycite{sil} & - \\ \hline 

\mycite{upadhyayjoint} & \url{https://github.com/shyamupa/xelms}  \\ \hline 

\mycite{cao2018} & \url{https://github.com/TaoMiner/NCEL} \\ \hline 

\mycite{deeptype} & {\url{https://github.com/openai/deeptype}}  \\ \hline 

\mycite{mueller} & \url{https://github.com/davidandym/wikilinks-ned}  \\ \hline

\mycite{shahbazi2019entity} & - \\ \hline 

\mycite{zero-shot} & \url{https://github.com/lajanugen/zeshel} \\ \hline 

\mycite{gillick} & \url{https://github.com/google-research/google-research/tree/master/dense_representations_for_entity_retrieval} \\ \hline 

\mycite{knowbert} & {\url{https://github.com/allenai/kb}} \\ \hline 

\mycite{titov1} & \url{https://github.com/lephong/dl4el} \\ \hline

\mycite{titov2} & \url{https://github.com/lephong/wnel} \\ \hline 

\mycite{reinforcement}  & -  \\ \hline 

\mycite{martins} & -  \\ \hline 

\mycite{yang2019} & \url{https://github.com/YoungXiyuan/DCA}  \\ \hline 

\mycite{ijcai2019-740} & \url{https://github.com/DeepLearnXMU/RRWEL} \\ \hline
 
\mycite{zhou-etal-2019-towards}&\url{https://github.com/shuyanzhou/burn_xel} \\ \hline

\mycite{broscheit2020investigating}& {\url{https://github.com/samuelbroscheit/entity_knowledge_in_bert} }\\ \hline 

\mycite{hou-etal-2020-improving} & \url{https://github.com/fhou80/EntEmb}\\ \hline

\mycite{onoe2020fine} &  \url{https://github.com/yasumasaonoe/ET4EL} \\ \hline

\mycite{chen-etal-2020-contextualized}& - \\ \hline

\mycite{wufacebook} & \url{https://github.com/facebookresearch/BLINK} \\ \hline 

\mycite{debayan}& {\url{https://github.com/debayan/pnel}} \\ \hline

\mycite{wu-dgcn}& \url{https://github.com/wujsAct/DGCN_EL} \\ \hline 

\mycite{fang-www} & \url{https://github.com/fangzheng123/SGEL}\\ \hline 

\mycite{chen2020improving}& - \\ \hline

\mycite{botha2020entity} & \url{http://goo.gle/mewsli-dataset} \\ \hline

\mycite{yao-etal-2020-zero} & \url{https://github.com/seasonyao/Zero-Shot-Entity-Linking} \\ \hline 

\mycite{li-etal-2020-efficient} & \url{https://github.com/facebookresearch/BLINK/tree/master/elq} \\ \hline 

\mycite{poerner-etal-2020-e} & \url{https://github.com/npoe/ebert} \\ \hline 

\mycite{fu2020design} & \url{http://cogcomp.org/page/publication_view/911} \\ \hline 

\mycite{mulang2020evaluating} & \url{https://github.com/mulangonando/Impact-of-KG-Context-on-ED} \\ \hline

\mycite{yamada20} & \url{https://github.com/studio-ousia/luke} \\  \hline 

\mycite{mrc}  & - \\  \hline 

\mycite{tang2021bidirectional} & - \\  \hline 

\mycite{de2020autoregressive}  &\url{https://github.com/facebookresearch/GENRE} \\  \hline 

\end{tabular}}

\label{table:sourcecodes}
\end{table*}

\end{appendix}
\twocolumn



\nocite{*} 
\bibliographystyle{ios1}           
\bibliography{bibliography}        

\end{document}